\documentclass{article}

\usepackage[preprint, nonatbib]{neurips_2026}

\usepackage[utf8]{inputenc}
\usepackage[T1]{fontenc}
\usepackage{hyperref}
\usepackage{url}
\usepackage{booktabs}
\usepackage{amsfonts}
\usepackage{amsmath}
\usepackage{nicefrac}
\usepackage{microtype}
\usepackage{xcolor}

\usepackage{graphicx}
\usepackage{multirow}
\usepackage{bm}
\usepackage{makecell}
\usepackage{xspace}
\usepackage{caption}
\usepackage[capitalize]{cleveref}
\usepackage{float}

\usepackage{eccvabbrv}
\usepackage{enumitem} 

\newcommand{\ours}{\textsc{Lasagna}}
\newcommand{\dataset}{\textsc{Lasagna-48K}}
\newcommand{\benchmark}{\textsc{LasagnaBench}}

\title{A Unified and Controllable Framework for Layered Image Generation with Visual Effects}

\author{
Jinrui Yang$^{1,2}$\thanks{This work was done when Jinrui Yang was a research intern at Adobe Research.} \quad
Qing Liu$^{2}$ \quad
Yijun Li$^{2}$ \quad
Mengwei Ren$^{2}$ \\
\bfseries
Letian Zhang$^{1}$ \quad
Zhe Lin$^{2}$ \quad
Cihang Xie$^{1}$ \quad
Yuyin Zhou$^{1}$ \\
\\
\normalfont
$^1$UC Santa Cruz \qquad
$^2$Adobe Research \\
\url{https://rayjryang.github.io/LASAGNA-Page/}
}

\begin{document}

\maketitle

\begin{figure}[!ht]
  \centering
  \includegraphics[width=1.0\linewidth]{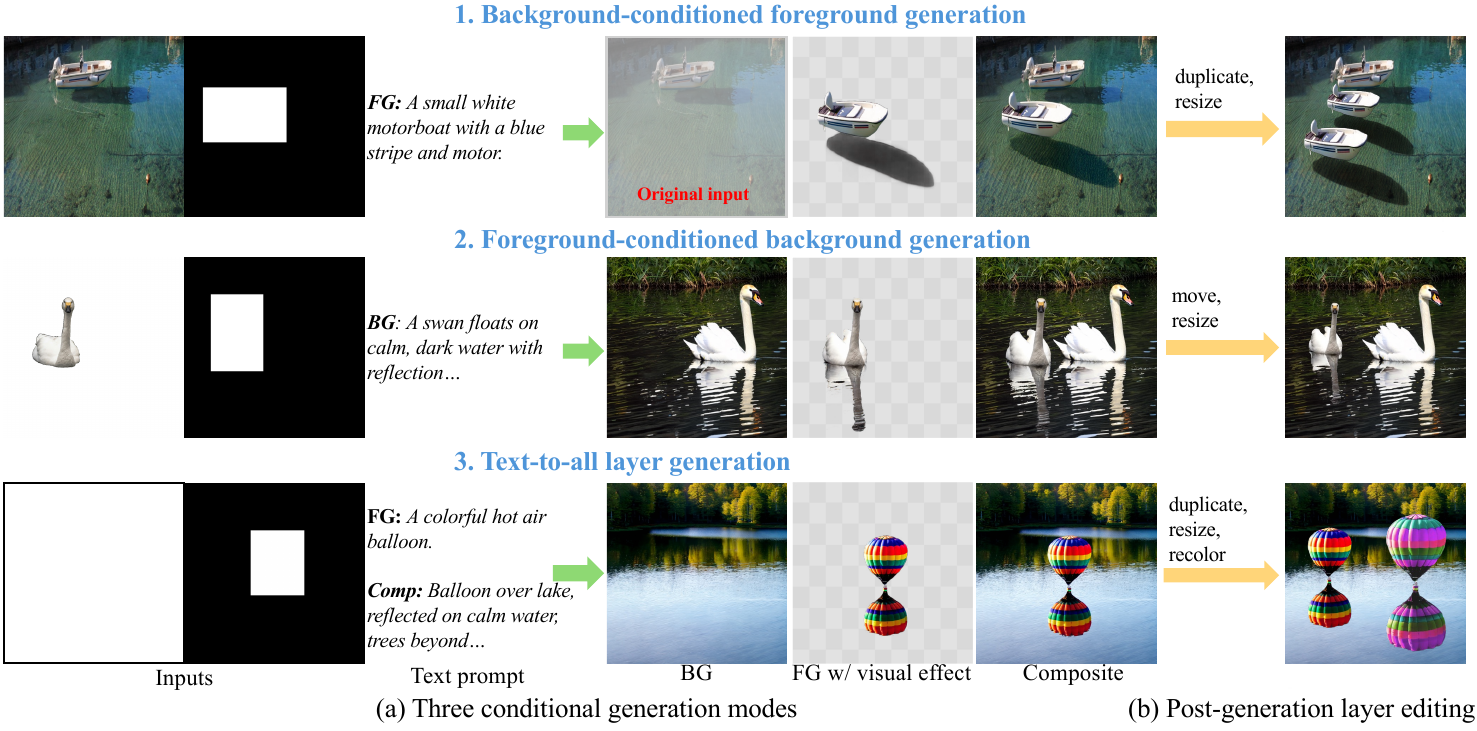}
  \caption{\textbf{Layered generation with \ours{}.}
    (a) Our framework supports three generation modes: background-conditioned foreground generation, foreground-conditioned background generation, and text-to-all layer generation, which flexibly handle different inputs and jointly synthesize coherent, high-quality composites, backgrounds, and transparent foregrounds with realistic visual effects (\eg, shadows and reflections).
    (b) Generated layers enable direct post-editing into new, coherent scenes. See the video demo in the supplementary material.}
  \label{fig:teaser_image}
\end{figure}

\begin{abstract}
Recent image generation models produce impressive composites, but often fail to preserve the identity of user-provided content when editing specific elements: the surrounding scene may shift, and even the edited object’s appearance can drift from the original. Layered representations offer a natural remedy—they allow users to independently manipulate individual elements—but existing layered methods typically produce transparent foregrounds without realistic visual effects such as shadows and reflections, forcing the use of a second harmonization model after every edit, which in turn reintroduces drift. To overcome these limitations, we present \ours{}, which generates a photorealistic background and an RGBA foreground with compelling visual effects in a single forward pass. By treating object-associated visual effects as part of the foreground layer, \ours{} supports the dominant class of consumer edits (e.g., translation, scaling, recoloring, duplication) via alpha compositing alone, without invoking any model post-edit, thereby eliminating the identity drift inherent to cascade editing pipelines. This single-pass design contrasts with prior layered methods that rely on separate expert models for each task; \ours{} handles diverse conditioning inputs—text prompts, foreground (FG), background (BG), and location masks—within a unified architecture. We further release two community resources: \dataset{}, the first public dataset of 48K layered image triplets with photorealistic visual effects (curated via a VLM-based filter trained on 30K human-labeled examples), and \benchmark{}, the first standardized benchmark for layer-centric generation and editing, comprising 242 expert-annotated samples across six diverse sources. Experiments show that \ours{} outperforms both general-purpose editors (FLUX, Qwen-Image-Edit, gpt-image-1) and prior layered methods (LayerDiffuse) across three generation modes, and supports a wide range of post-edits without any model re-inference.
\end{abstract}

\section{Introduction}
\label{sec:intro}

Recent advances in text-to-image generation have predominantly leveraged diffusion-based generative models, enabling impressive synthesis quality and semantic accuracy from text prompts~\cite{rombach2022high,ramesh2021zero,podell2023sdxl,esser2024scaling,labs2025flux,wu2025qwen}. Despite their success, these models typically produce images as a single entity, limiting controllability for real-world editing tasks. Consequently, modifications to individual elements within a generated image---such as repositioning, scaling, or adjusting a specific object---often require complex prompt engineering or re-generating the entire image, making it difficult to preserve desired attributes in other regions.

To achieve controllable editing, recent works~\cite{zhong2025tp,zhang2024transparent,dalva2024layerfusion,huang2025psdiffusion} explore compositional and layered image generation. This approach, which decomposes generated images into layers, allows for independent manipulation of image components. However, current layered approaches fall short in several critical aspects that prevent their use in real-world scenarios:
\begin{enumerate}[
    label=\Roman*.,
    leftmargin=*,
    labelsep=0.5em,
    align=left,
    noitemsep,
    topsep=0pt,
    partopsep=0pt,
    parsep=0pt
]
    \item \textbf{Lack of visual effects in foreground layers.} The faithful generation of visual effects like shadows and reflections intrinsically associated with the foreground object is largely overlooked.
    \item \textbf{Lack of a unified, generative layer framework.} Current approaches often lack a unified framework capable of handling diverse conditional inputs such as foreground (FG), background (BG), masks, and text, thereby limiting their controllability and practical utility.
    \item \textbf{Absence of public data and standardized evaluation.} As shown in \cref{tab:overall_papers}, most existing methods rely on proprietary or non-public training data. Public datasets like MULAN~\cite{tudosiu2024mulan} still fall short, as they lack realistic FG visual effects essential for downstream editing. Moreover, the absence of standardized evaluation protocols further hinders meaningful progress comparison across studies.
\end{enumerate}

In this work, we address these limitations and enable controllable, versatile, and realistic layered editing from three complementary perspectives: a unified generation paradigm, publicly available training data, and a standardized benchmark. We present \ours{}, a novel framework designed to generate images as a composition of FG layers and background layers, explicitly embedding visual effects such as shadows and reflections. Unlike standard editing models that output a single flattened composite, our layered representation guarantees that unedited content is perfectly preserved while remaining freely editable for downstream modifications. This unified architecture simultaneously integrates diverse conditioning inputs and supports three generation modes: background-conditioned foreground layer generation (\textbf{FG\_Gen}), foreground-conditioned background generation (\textbf{BG\_Gen}), and text-to-all layer generation (\textbf{Text2All}). In \textbf{BG\_Gen}, where user-provided foreground assets typically lack effects, \ours{} restores the missing effects while preserving identity, producing an editable RGBA FG for subsequent operations.

To enable \ours{} training, we introduce \dataset{}, the first publicly available dataset of 48K natural images with faithfully decomposed RGBA FG and BG layers. Critically, these FGs accurately preserve effects like shadows and reflections in relation to the object and its transparency.
We will release \dataset{} to facilitate the research and development of models capable of capturing these complex, scene-consistent visual effects.

Furthermore, we introduce \benchmark{} to establish a standardized measure for our method and future research. Evaluation in layer editing and generation has been challenging, as prior work relies on bespoke protocols and user studies. \benchmark{} provides the first public benchmark for this task, featuring 242 real-world images sourced from 6 diverse datasets, each meticulously decomposed by human experts into high-fidelity, text-paired layers that accurately capture complex visual effects. On \benchmark{}, our method achieves superior layer generation while preserving object identity, spatial fidelity, and visual coherence.

In summary, our primary contributions are:
\textbf{(1)} We present \textbf{\ours{}}, a unified framework supporting three generation modes and flexible conditioning inputs (text, images, masks). It synthesizes realistic composites by jointly or individually generating coherent BGs and RGBA FGs with visual effects, enabling highly controllable, professional photo-editing-tool-style image editing without extra inference.
\textbf{(2)} A new dataset \textbf{\dataset{}}, the first publicly available dataset featuring over 48K natural images with decomposed BG layers as well as FG layers with scene-coherent visual effects.
\textbf{(3)} \textbf{\benchmark{}}, the first public benchmark for rigorous and standardized evaluation of controllable layer-centric generation and editing.
\textbf{(4)} Ours achieves high-quality layer generation and editing, particularly for tasks requiring strict identity preservation and harmonious integration.

\section{Related Works}
\label{sec:related_work}

\begin{table*}[t]
    \centering
    \footnotesize
    \renewcommand{\arraystretch}{1.03}

    \begin{minipage}{0.54\textwidth}
        \centering
        \setlength\tabcolsep{1.5pt}

        \captionof{table}{
        \textbf{Layer Dataset Overview.}
        $\checkmark$: available; $\times$: not available or non-public.
        }

        \resizebox{\textwidth}{!}{
        \begin{tabular}{l c c c}
            \toprule
            \textbf{Paper} & \textbf{Layer data (Public)} & \textbf{Eval Bench } & \textbf{Visual Effect} \\
            \midrule
            MULAN~\cite{tudosiu2024mulan} & $\checkmark$ ($\checkmark$) & $\times$ & $\times$ \\
            LayerDiffuse~\cite{zhang2024transparent} & $\checkmark$ ($\times$) & $\times$ & $\times$ \\
            PSDiffusion~\cite{huang2025psdiffusion} & $\checkmark$ ($\times$) & $\times$ & $\times$ \\
            \midrule
            \textbf{\ours{} (ours)} & $\checkmark$ ($\checkmark$) & $\checkmark$ & $\checkmark$ \\
            \bottomrule
        \end{tabular}
        }
         \vspace{-7pt}
        \label{tab:overall_papers}
    \end{minipage}
    \hspace{0.02\textwidth}
    \begin{minipage}{0.42\textwidth}
        \centering
        \setlength\tabcolsep{1.0pt}

        \captionof{table}{
        \textbf{Overview of the three generation modes.}
        }

        \resizebox{\textwidth}{!}{
        \begin{tabular}{l l l}
            \toprule
            \textbf{Mode} & \textbf{Inputs} & \textbf{Targets} \\
            \midrule
            FG\_Gen & $\{\mathbf{c}_\text{txt}, \mathbf{c}_\text{mask}, \mathbf{c}_\text{bg}\}$ 
                    & $\{\mathbf{x}_0^\text{comp}, \mathbf{x}_0^\text{fg+ve}\}$ \\
            BG\_Gen & $\{\mathbf{c}_\text{txt}, \mathbf{c}_\text{mask}, \mathbf{c}_\text{fg}\}$ 
                    & $\{\mathbf{x}_0^\text{comp}, \mathbf{x}_0^\text{bg}, \mathbf{x}_0^\text{fg+ve}\}$ \\
            Text2All & $\{\mathbf{c}_\text{txt}, \mathbf{c}_\text{mask}\}$ 
                     & $\{\mathbf{x}_0^\text{comp}, \mathbf{x}_0^\text{bg}, \mathbf{x}_0^\text{fg+ve}\}$ \\
            \bottomrule
        \end{tabular}
        }

        \label{tab:modes}
    \end{minipage}

\end{table*}

\subsection{Text-to-Image and Image Editing Models}

Recent text-to-image diffusion models~\cite{rombach2022high,ramesh2021zero,podell2023sdxl,esser2024scaling,labs2025flux,wu2025qwen} have made remarkable progress in generating high-fidelity images from text. However, these models are typically confined to single-layer synthesis, lacking an explicit layered representation. Consequently, they cannot produce RGBA outputs or support independent post-generation editing of specific elements without unintended changes to other regions. While specialized image editing models~\cite{labs2025flux,wu2025qwen,chen2025unireal,zhong2025tp,zhang2023adding,zhao2024ultraedit,liu2025step1x} have been developed for common editing tasks, they still struggle with precise object-level control and often introduce non-local artifacts. They are particularly weak in complex spatial edits, such as enlarging or relocating an object while preserving its identity and appearance, as they lack an understanding of the layered composition of the scene. This motivates the development of layer-centric frameworks that inherently support structured, controllable synthesis and editing.

\subsection{Image Layer Generation}

To enable compositional editing, prior work has explored two main paradigms for layered generation.

\noindent\textbf{Image-layer extraction via post-processing:} This common pipeline first uses text-to-image models~\cite{rombach2022high,ramesh2021zero,podell2023sdxl,esser2024scaling,labs2025flux,wu2025qwen} to generate an RGB composite image. Then, existing segmentation models~\cite{ren2024grounded,ravi2024sam,kirillov2023segment} can be used to extract an independent FG layer. Finally, inpainting models (\eg,~\cite{zhao2025objectclear,wei2025omnieraser,winter2024objectdrop,zhuang2024task,ju2024brushnet}) are typically employed to reconstruct the occluded BG. However, this multi-stage, separately optimized pipeline accumulates errors across stages and often fails to preserve global coherence and cross-layer consistency during post-editing. Operations such as object translation or scaling often produce spatially inconsistent or visually unnatural results.

\noindent\textbf{Direct transparent image layer generation:} This paradigm aims to generate layers directly. While initial methods such as those of Fontanella et al.~\cite{fontanella2024generating} and LayerDiffuse~\cite{zhang2024transparent} target simple scenes, recent works like DreamLayer~\cite{huang2025dreamlayer} and PSDiffusion~\cite{huang2025psdiffusion} extend this to multi-layer synthesis. However, these models predominantly focus on cartoon or synthetic domains and lack complex physical effects (e.g., shadows and reflections). Furthermore, they lack conditional synthesis capabilities (e.g., BG\_Gen or FG\_Gen) and explicit spatial control, leading to uncontrollable object placement that fails to meet professional editing standards. In design-centric tasks, ART~\cite{pu2025art} utilizes layout conditioning but remains limited in editing versatility and realistic effect modeling. Other approaches, such as LayerDecomp~\cite{yang2025generative} and Qwen-Image-Layered~\cite{yin2025qwen}, focus on layer decomposition but cannot generate novel content. Crucially, none of these methods provide a unified, controllable framework for generating transparent FGs with high-fidelity, physically-grounded visual effects. Unlike prior works trained on synthetic or design-centric data, \ours{} leverages real-world images (e.g., COCO) and supports three generation modes with mask-guided spatial constraints, effectively addressing the key barriers to realistic, professional-grade image editing.

\subsection{Layer Dataset}
\vspace{-8pt}

Previous studies~\cite{zhang2023text2layer,tudosiu2024mulan,kang2025layeringdiff, huang2025psdiffusion} have introduced several layer-related datasets.
MULAN~\cite{tudosiu2024mulan}, a prominent multi-layer dataset, provides object-level decompositions but does not include explicit visual effects as part of the FG layers. Text2Layer~\cite{zhang2023text2layer} generates a two-layer decomposition, but the dataset is not public and does not incorporate visual effects.
PSDiffusion~\cite{huang2025psdiffusion} proposes an internal multi-layer dataset, consisting of 30K samples. Except for MULAN, most datasets remain private and none explicitly account for visual effects. To bridge this data gap, we introduce \dataset{}, a new dataset built upon an advanced decomposition model that jointly generates BG and FG layers while faithfully preserving complex visual effects in the FG's alpha channel. In addition, we manually annotate a high-quality layer benchmark. All training and testing data are fully released to encourage transparency and foster further research in this area.

\section{Approach}
\label{sec:approach}
\vspace{-8pt}

\begin{figure*}[t]
  \centering
    \includegraphics[width=1.0\linewidth]{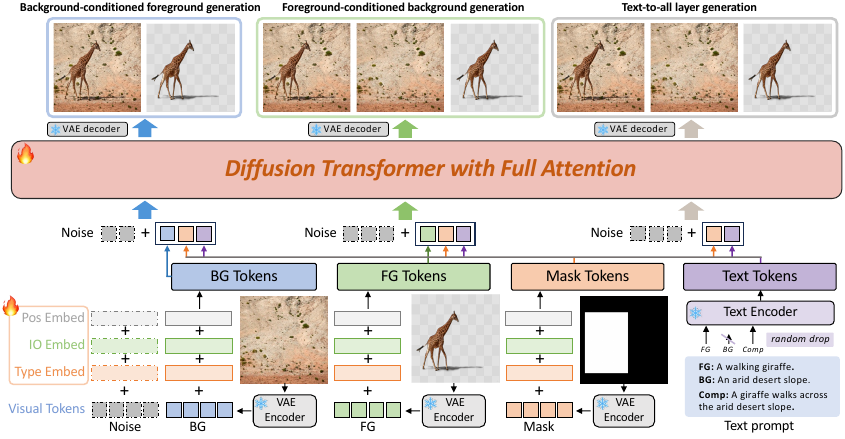}
   \caption{\textbf{Framework of \ours{}.} We formulate the joint generation of composite images, BGs, and FGs as a flexible, layer-conditional denoising task. This unified framework supports multiple workflows, including FG\_Gen, BG\_Gen, and Text2All. We use a unified input with learnable embeddings to distinguish roles of visual latents (noise, BG, FG, and mask) across tasks, enabling the model to adapt its behavior under various generation settings. This allows a single attention-based model to flexibly process varied combinations of inputs and targets simultaneously.}
   \label{fig:framework}
\end{figure*}

\subsection{\ours{} Framework}
\vspace{-5pt}

As shown in \cref{fig:framework}, \ours{} models the joint generation of composite images $\mathbf{x}^\text{comp}$, BGs $\mathbf{x}^\text{bg}$, and
FGs with visual effects $\mathbf{x}^\text{fg+ve}$ as a flexible, layer-conditional denoising task.
Our model learns to denoise a set of target images
$\mathbf{X}_t \subseteq \{\mathbf{x}_t^\text{comp}, \mathbf{x}_t^\text{bg}, \mathbf{x}_t^\text{fg+ve}\}$  conditioned on a set of inputs
$\mathbf{C} \subseteq \{\mathbf{c}_\text{txt}, \mathbf{c}_\text{mask}, \mathbf{c}_\text{bg}, \mathbf{c}_\text{fg}\}$. By varying the composition of $\mathbf{X}_t$ and $\mathbf{C}$, we unify three generation modes (FG\_Gen, BG\_Gen, and Text2All) in a single model, addressing different real-world editing needs (see \cref{tab:modes}).

We build \ours{} upon the Diffusion Transformer (DiT) architecture~\cite{peebles2023scalable, chen2025unireal} to support flexible editing tasks, adapted to handle heterogeneous inputs. We employ four embedding types to distinguish between different tasks and image types:
\begin{itemize}[noitemsep, topsep=0pt, partopsep=0pt, parsep=0pt]
    \item \textit{Type Embedding}---represents the semantic role of each image, \eg, BG or FG.
    \item \textit{IO Embedding}---indicates if a frame is used as an input or an output in the current task.
    \item \textit{Position Embedding}---joint spatial and frame position of image tokens.
    \item \textit{Timestep Embedding}---the diffusion step.
\end{itemize}
Text prompts are encoded by T5~\cite{raffel2020exploring}. All conditional tokens and noisy target tokens are concatenated into a single sequence, allowing the model's self-attention blocks to seamlessly integrate information from any arbitrary set of conditions $\mathbf{C}$ to guide the denoising of targets $\mathbf{X}_t$.

We train our model by optimizing a unified denoising objective across all conditional generation tasks. Each mode uses its specific conditional inputs $\mathbf{C}^{(m)}$ and targets $\mathbf{X}_0^{(m)}$ as defined in \cref{tab:modes}. The model is trained to minimize the joint expectation $\mathcal{L}_{\mathrm{dm}}$ over this multi-task distribution:
\begin{align*}
    \mathcal{L}_{\mathrm{dm}} = \mathbb{E}_{m, t, \mathbf{X}_0^{(m)}, \boldsymbol{\epsilon}} \left[ \| \boldsymbol{\epsilon}_\theta(\mathbf{X}_t^{(m)}; \mathbf{C}^{(m)}, t) - \boldsymbol{\epsilon} \|_2^2\right] \\
    s.t.~~ \mathbf{X}_t^{(m)} = \sqrt{\alpha_t} \mathbf{X}_0^{(m)} + \sqrt{1 -\alpha_t} \boldsymbol{\epsilon}, \text{ and } \boldsymbol{\epsilon} \sim \mathcal{N}(\boldsymbol{0}, I). \notag
\end{align*}
The training loss follows flow matching~\cite{lipman2022flow}. See more technical details in \cref{app:framework_details} in the Appendix.

\vspace{-10pt}
\subsection{\dataset{} Dataset}
\label{sec:LASAGNA-48K}
\vspace{-8pt}

To enable the training of our controllable, layer-based framework, we introduce \dataset{}, the first publicly available large-scale dataset of over 48K high-quality image triplets with visual effects.

\vspace{-10pt}

\paragraph{Data sources.}
We build our dataset from three public sources: \textsc{MULAN}~\cite{tudosiu2024mulan}, \textsc{COCO 2017}~\cite{lin2014microsoft}, and \textsc{SOBA}~\cite{wang2022instance}. MULAN provides layered data but lacks visual effects. COCO offers diverse scenes with complex layouts for robust editing. SOBA, a shadow dataset, adds rich visual effects for realistic composites. Finally, \dataset{} includes 8K MULAN, 39K COCO, and 1K SOBA samples.

\vspace{-5pt}

\begin{figure}[t]
  \centering
   \includegraphics[width=0.7\linewidth]{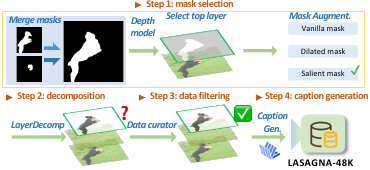}
   \caption{\textbf{Data construction pipeline.} Starting with existing datasets, we implement a four-stage data construction pipeline leveraging off-the-shelf models with a custom-trained data curator. This process yields a high-quality dataset as the foundation for subsequent model training.}
   \label{fig:data_curator}
\end{figure}

\begin{figure*}[t]
  \centering
    \includegraphics[width=0.9\linewidth]{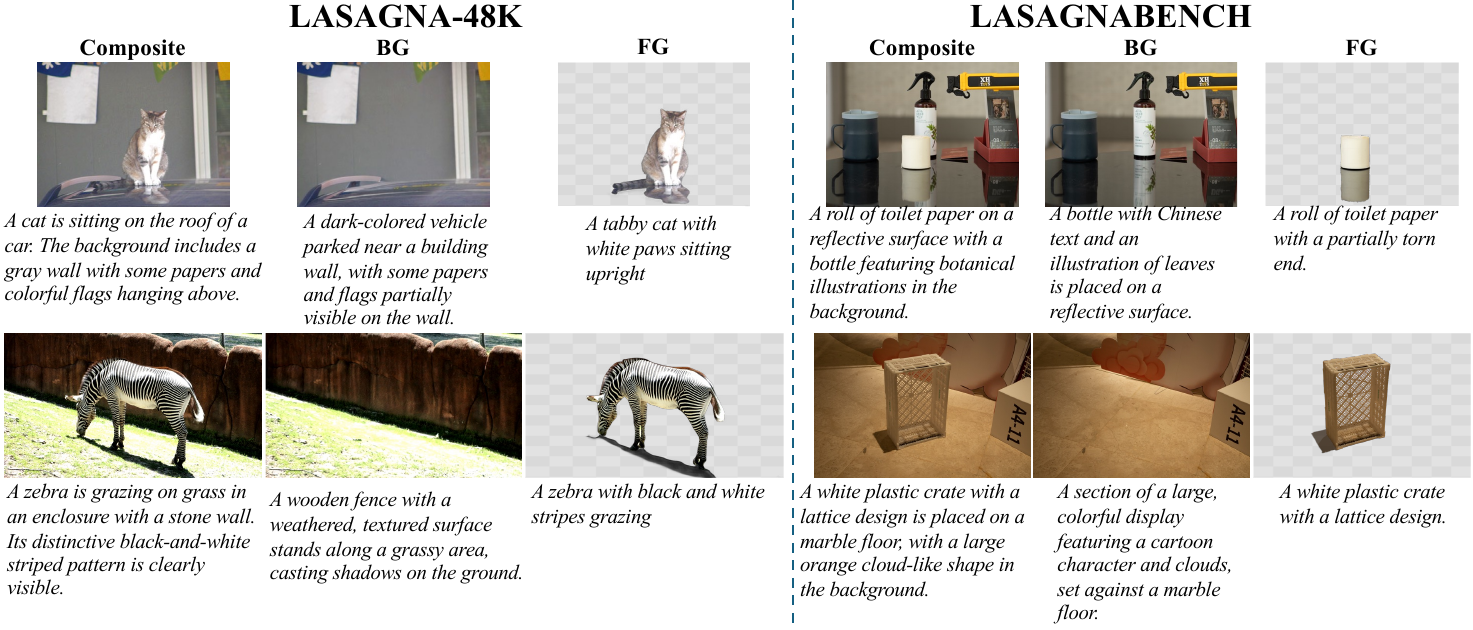}
   \caption{\textbf{Samples of \dataset{} and \benchmark{}.} Each  sample consists of a composite image, a clean BG, and a FG layer with visual effects, along with corresponding captions for all components. Additional examples are provided in \cref{sec:more_samples_dataset} in the Appendix.}
   \label{fig:vis_train_and_bench}
   \vspace{-1em}
\end{figure*}

\paragraph{Data construction pipeline.}
We design a four-stage pipeline to ensure high data fidelity (\cref{fig:data_curator}); examples of generated triplets and captions are shown in \cref{fig:vis_train_and_bench}.
\begin{enumerate}[
    label=\Roman*.,
    leftmargin=*,
    labelsep=0.5em,
    align=left,
    noitemsep,
    topsep=0pt,
    partopsep=0pt,
    parsep=0pt
]
    \item \emph{Non-occluded mask selection.} To isolate unoccluded foremost objects without direct annotations, we generate multiple mask proposals per object utilizing available dataset-specific annotations. Depending on the source, we extract parallel base masks using explicit layer annotations, depth-filtered instance masks~\cite{ke2025marigold}, and a salient segmentation model~\cite{gao2024multi}. To handle boundary inaccuracies, we augment each base mask with a dilated variant. All proposals are processed independently, and our data curator (Step III) selects the highest-scoring outcome as the final clean layer.
    \item \emph{LayerDecomp decomposition.} We process each image and its mask variants with LayerDecomp~\cite{yang2025generative}, extracting multiple sets of BGs and FGs with visual effects.
    \item \emph{Data filtering.} Due to the consistency loss in LayerDecomp, the quality of extracted BGs and FGs is inherently correlated — enabling BG fidelity alone as a reliable proxy for overall quality. We train a data curator built upon InternVL2.5-8B~\cite{chen2024expanding} to assess the BGs fidelity and filter low-quality data. The curator is trained on $30$K carefully human-annotated samples (see Appendix), and achieves $88.8\%$/$72.3\%$ precision/recall on a 1K human-annotated held-out test set. After filtering, we further use Qwen2.5-VL-32B~\cite{bai2025qwen2} to remove residual artifacts in FG. Finally, a manual inspection of 200 random \dataset{} samples shows that \textgreater90\% are judged to be good.
    \item \emph{Captioning.} After obtaining high-quality triplets of composite images, BGs, and FGs, we prompt InternVL 2.5-38B~\cite{chen2024expanding} to caption these images jointly, considering cross-image relationships for semantically consistent descriptions.
\end{enumerate}

\vspace{-8pt}
\section{Experiments}
\label{sec:experiment}
\vspace{-8pt}

\paragraph{\ours{} Implementation.} We finetune a pre-trained 2B-parameter DiT model. All conditional images and targets are encoded with RGBA-VAE, which is finetuned from DiT VAE using a combination of L1, GAN, and perceptual losses~\cite{zhang2018unreasonable}.
We adopt resolution-specific batching (batch size $6$ for $512^2$ resolution, 1 for $1024^2$) to improve generalization across scales. We use the AdamW optimizer and set the learning rate to $1.2\times10^{-5}$ with linear warm-up for $2K$ steps. The model is trained for $20\text{K}$ iterations. For inference, results are generated using DDIM sampling for $50$ steps.

\subsection{Benchmark}
\vspace{-5pt}

We introduce \benchmark{}, the first public benchmark specifically designed for layer-centric image generation. Prior works~\cite{dalva2024layerfusion,zhang2024transparent,huang2025psdiffusion} use varied protocols on non-public datasets. \benchmark{} contains 242 samples from 6 diverse datasets: 4 public sources~\cite{tudosiu2024mulan,lin2014microsoft,wang2022instance,unsplash} and 2 in-house data sources, all of which will be released (see \cref{fig:vis_train_and_bench} and more benchmark details in Table~\ref{tab:benchmark_distri} in the Appendix). Each sample includes a real photographic composite image, a BG image, an FG with visual effects annotated by professional annotators, and captions.
For public datasets, BGs were generated with LayerDecomp, followed by automated curation and manual verification to ensure consistency. For in-house data, we captured controlled real-world photography pairs, following a similar data collection methodology as in prior works~\cite{yang2025generative,winter2024objectdrop}. These collection efforts ensure that \benchmark{} offers a diverse and high-quality set of layer representations, capturing realistic variations in object appearance and surrounding BG scenes, along with visually authentic visual effects.

\subsection{Layer Generation vs General Models}
\vspace{-8pt}

\begin{figure*}[t]
  \centering
   \includegraphics[width=1.0\linewidth]{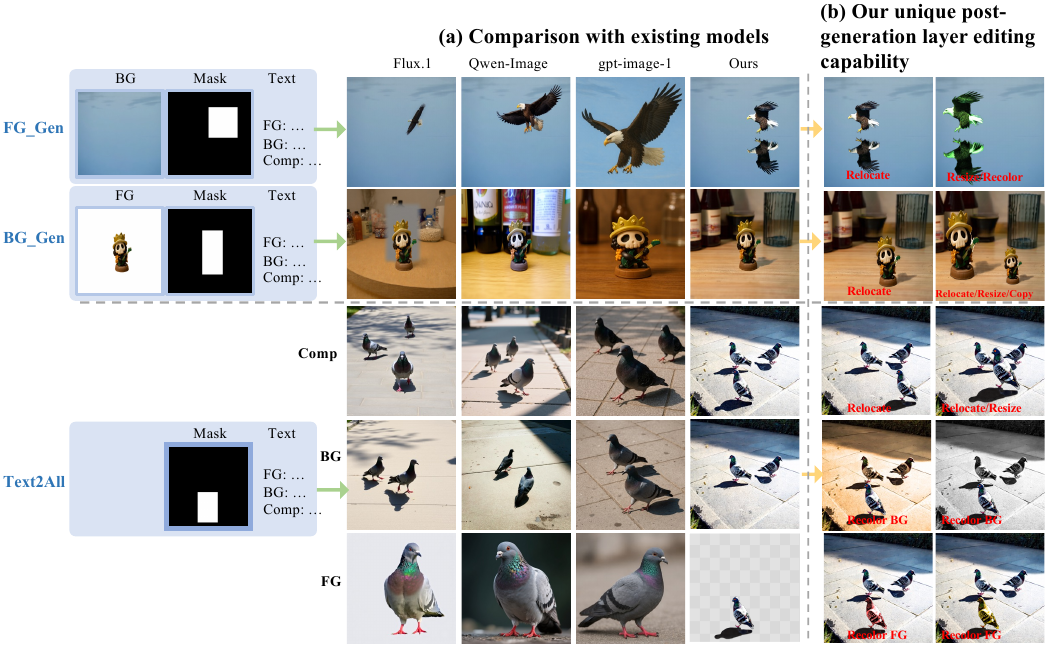}
   \vspace{-14pt}
    \caption{\textbf{Layer generation compared with state-of-the-art image generation and editing models.} We compare \ours{} with Flux.1~\cite{labs2025flux,huggingface2025flux}, Qwen-Image-Edit~\cite{wu2025qwen}, and gpt-image-1[High]~\cite{openai_gptimage1_2025}. \textbf{(a)} Across three distinct generation tasks, \ours{} consistently achieves superior inter-layer coherence and consistency. In contrast, competing models often fail to maintain these properties. \textbf{(b)} Moreover, by generating FGs with faithfully preserved visual effects, \ours{} enables diverse post-generation editing operations on individual layers directly---a capability not supported by existing models. See more visualization results in \cref{sec:more_model_results} in the Appendix.}
   \label{fig:com_others}
\end{figure*}

We compare \ours{} with three leading image generation and editing models: FLUX.1~\cite{labs2025flux,huggingface2025flux}, Qwen-Image~\cite{wu2025qwen}, and gpt-image-1[high]~\cite{openai_gptimage1_2025}.
Since these models do not natively support multi-task generation, we employ their task-specific variants (\eg, inpainting/editing models for conditional tasks, T2I models for Text2All) to ensure fair comparison at their best performance. Because these baselines also lack multi-layer generation capability, we focus evaluation on the generated composite images.
We employ FID~\cite{heusel2017gans,parmar2021cleanfid} and CLIP-FID~\cite{parmar2021cleanfid} to measure image quality and semantic alignment. Following Complex-Edit~\cite{yang2025complexedit}, we also use a GPT-4o-based score to evaluate Instruction Following and Identity Preservation, averaging them into a final score of 0 to 10.

As shown in \cref{tab:quant:clipfid_fid}, \ours{} consistently outperforms prior methods across all modes, achieving higher image quality and better semantic alignment with the input instructions.
Qualitative results in \cref{fig:com_others} show superior inter-layer coherence and faithful visual effects.
In contrast, competing models generate inconsistent compositions or misalign FGs and BGs. By generating environment-aware visual effects in FG layers, \ours{} enables direct post-editing and recomposition, which are not supported by existing methods. Beyond our benchmark, we also report quantitative comparisons on the public benchmarks ImgEdit-Bench~\cite{ye2025imgedit} and GenEval~\cite{ghosh2023geneval}. Detailed results are provided in the supplementary material.

\begin{table*}[t]
    \setlength\tabcolsep{4pt}
    \renewcommand{\arraystretch}{1.1}
    \centering

    \caption{\textbf{Comparison with general models for layer generation.}
    Results for models marked with $\ast$ are obtained using their respective expert models rather than a single unified model. Specifically, for the FG\_Gen and BG\_Gen tasks, we use the FLUX.1-Fill-dev, Qwen-Image-Edit-2509, and gpt-image-1[high] editing models, respectively. For the Text2All task, we use the FLUX.1-schnell, Qwen-Image, and gpt-image-1[high] models as text-to-image models, respectively.}
    \resizebox{\textwidth}{!}{%
    \begin{tabular}{l c cccc cccc cccc}
        \toprule
        \multirow{2}{*}{Model} & \multirow{2}{*}{\# Params}
        & \multicolumn{4}{c}{FG\_Gen}
        & \multicolumn{4}{c}{BG\_Gen}
        & \multicolumn{4}{c}{Text2All} \\
        \cmidrule(lr){3-6}\cmidrule(lr){7-10}\cmidrule(lr){11-14}
        & & Cond Image & CFID $\downarrow$ & FID $\downarrow$ & GPT Score $\uparrow$
          & Cond Image & CFID $\downarrow$ & FID $\downarrow$ & GPT Score $\uparrow$
          & Cond Image & CFID $\downarrow$ & FID $\downarrow$ & GPT Score $\uparrow$ \\
        \midrule
        gpt-image-1[high]\textsuperscript{$\ast$}               & --- & BG & 20.3 & 116.9 & 8.8 & FG & 20.3 & 115.2 & 8.9 & None & 25.6 & 130.8 & 7.1 \\
        FLUX.1\textsuperscript{$\ast$}        & 12B & BG, Mask & 10.0 & 79.6  & 8.9 & FG, Mask & 16.1 & 105.9 & 8.5 & None & 22.7 & 131.1 & 6.0 \\
        Qwen-Image-Edit\textsuperscript{$\ast$} & 20B & BG, Mask & 12.1 & 92.5  & 9.0 & FG, Mask & 15.1 & 101.2 & 7.9 & None & 24.9 & 131.5 & 5.7 \\
        \textbf{Ours}                              & 2B  & BG, Mask & \textbf{9.7}  & \textbf{72.0}  & \textbf{9.3} & FG, Mask & \textbf{14.1} & \textbf{98.6}  & \textbf{9.0} & Mask & \textbf{16.9} & \textbf{115.8} & \textbf{7.6} \\
        \bottomrule
    \end{tabular}%
    }
    \vspace{-10pt}
    \label{tab:quant:clipfid_fid}
\end{table*}

\subsection{Layer Generation vs Expert Model}
\vspace{-8pt}

\begin{figure*}[t]
  \centering
    \includegraphics[width=0.95\linewidth]{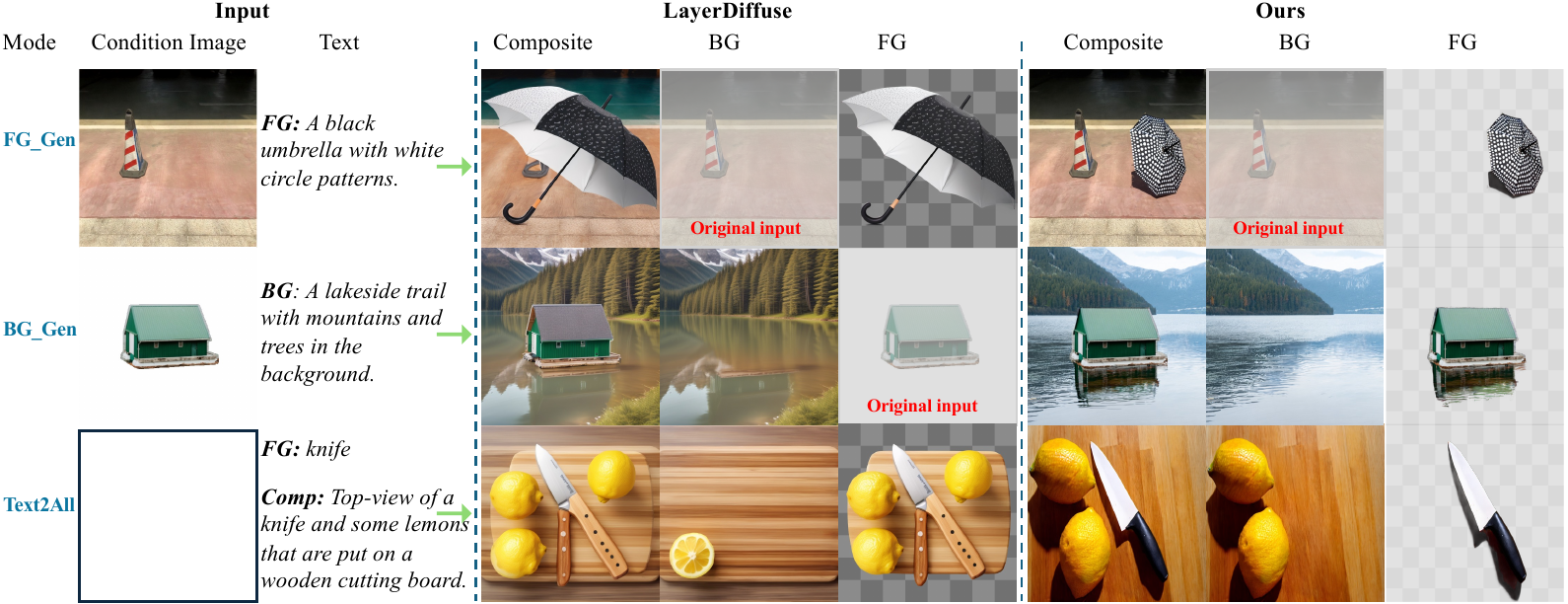}
   \caption{\textbf{Layer generation compared with LayerDiffuse~\cite{zhang2024transparent}.} In FG\_Gen, our model produces objects with appropriate size and position, along with realistic shadows consistent with the BG. In BG\_Gen and Text2All, our model produces visually consistent results across all layers. Furthermore, it can generate new FGs with corresponding visual effects, enabling flexible and realistic post-editing.}
   \label{fig:vis_comp_layerdiff}
\end{figure*}

We further compare \ours{} with LayerDiffuse~\cite{zhang2024transparent}, a prior work specifically designed for layered image generation through multiple expert modes and the only closely related public model. However, LayerDiffuse relies on separate, independently trained models for each task, limiting its controllability and consistency across layers. As shown in \cref{tab:comp_layerdiffuse}, \ours{} significantly outperforms LayerDiffuse across all three generation modes in terms of CLIP-FID, confirming the effectiveness of our unified framework in producing coherent and semantically faithful results across diverse generation settings. Qualitative comparisons in \cref{fig:vis_comp_layerdiff} highlight these improvements. FGs generated by LayerDiffuse often appear centered and lack positional diversity due to the absence of explicit spatial constraints. The model also frequently fails to clearly separate FG and BG regions and struggles to represent all described entities when given complex captions.
In contrast, \ours{} produces spatially controlled, semantically complete, and visually coherent results across all generation modes.

\vspace{-10pt}

\subsection{The Necessity of Layered Generation with Visual Effects}
\label{sec:ve_necessity}
\vspace{-6pt}
To quantify the value of explicit layer representations with visual effects---and to directly address whether naively cascading existing state-of-the-art models could achieve comparable results---we compare three editing paradigms across three representative editing tasks: recoloring, which alters static appearance attributes; spatial editing, which manipulates object position or scale (\eg, movement, resizing); and complex editing, which combines both to test multi-factor control. (see more details in \cref{sec:ve_necessity_supp} in the Appendix). The baselines use Qwen-Image (T2I) for composite synthesis and Qwen-Image-Edit-2509 for editing:

\begin{enumerate}[label=\Roman*.,
    leftmargin=*,
    labelsep=0.5em,
    align=left,
    noitemsep,
    topsep=0pt,
    partopsep=0pt,
    parsep=0pt]
\item \emph{Instruct Editing}: Feeds the Qwen-Image composite into Qwen-Image-Edit-2509 and performs direct text-guided modifications.
\item \emph{Cascaded Layer Editing}: A multi-model assembly chaining Qwen-Image $\rightarrow$ segmentation~\cite{liu2025seg} (FG extraction) $\rightarrow$ inpainting (BG restoration) $\rightarrow$ re-composition.
\item \emph{Layer Editing with Visual Effects (Ours)}: Perform the same explicit object-layer editing as above, while additionally incorporating the visual effects generated by \ours{}.
\end{enumerate}

\begin{figure}[t]
  \centering
\includegraphics[width=0.8\linewidth]{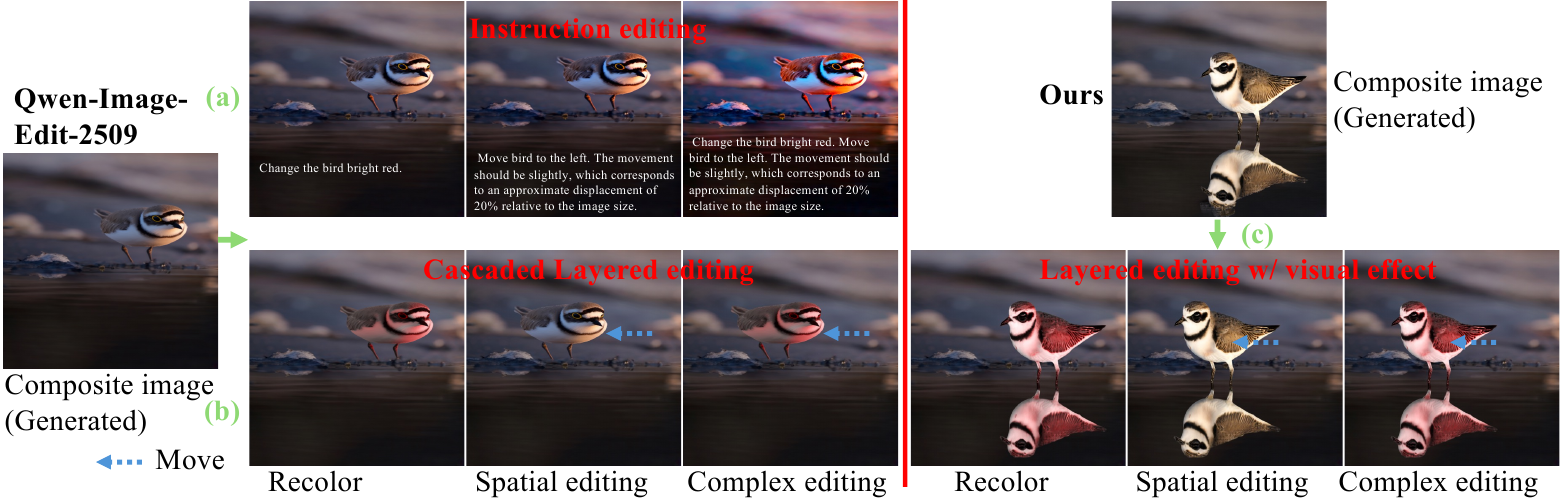}
  \caption{\textbf{The Necessity of Layered Generation with Visual Effects.} We demonstrate the benefits of explicit layer representations with visual effects by progressively comparing three paradigms: \emph{Instruct Editing}, \emph{Cascaded Layer Editing}, and \emph{Layer Editing with Visual Effects} (\ours{}). Across recoloring, spatial, and compositional editing tasks, the lack of explicit layer representations makes \emph{Instruct Editing} prone to unintended changes and less responsive to spatial instructions, while \emph{Layer Editing with Visual Effects}  yields more coherent and photorealistic results. \textit{Zoom in for details}.}
   \vspace{-10pt}
  \label{fig:vis_layer_editing}
\end{figure}

As shown in \cref{tab:quant:editing}, \emph{Cascaded Layer editing} already outperforms \emph{Instruct editing} across all editing goals, achieving higher fidelity and spatial consistency.
\emph{Layer editing with visual effects} achieves the best overall performance with substantial gains in perceptual realism and physical plausibility. Qualitative comparisons in \cref{fig:vis_layer_editing} further highlight these differences.
\emph{Instruct editing} often introduces unintended global changes (\eg, altered background tones during recoloring) and struggles to achieve precise spatial edits.
In contrast, \emph{Cascaded Layer editing} enables fine-grained spatial manipulation while preserving object identity, which is critical for user-driven image editing applications.
Finally, \emph{Layer editing with visual effects} produces the most visually coherent and realistic results, generating shadows and reflections that are physically consistent with the surrounding scene. These results confirm that explicitly modeling layered representations with physically grounded visual effects is crucial for achieving realistic, consistent, and controllable image manipulation. 

\begin{table*}[t]
    \centering
    \footnotesize
    \setlength\tabcolsep{2pt}
    \renewcommand{\arraystretch}{1.05}

    \begin{minipage}{0.49\textwidth}
        \centering
        \captionof{table}{
        \textbf{Comparison with LayerDiffuse~\cite{zhang2024transparent} for layer generation.}
        Each cell reports CLIP-FID($\downarrow$).
        }
        \vspace{-6pt}

        \resizebox{\textwidth}{!}{
        \begin{tabular}{l cc cc ccc}
            \toprule
            \multirow{2}{*}{Model}
            & \multicolumn{2}{c}{FG\_Gen}
            & \multicolumn{2}{c}{BG\_Gen}
            & \multicolumn{3}{c}{Text2All} \\
            \cmidrule(lr){2-3}\cmidrule(lr){4-5}\cmidrule(lr){6-8}
            & Comp. & FG
            & Comp. & BG
            & FG & Comp. & BG \\
            \midrule
            LayerDiffuse~\cite{zhang2024transparent}
            & 42.0 & 43.8
            & 43.2 & 43.1
            & 45.2 & 46.0 & 48.2 \\
            \textbf{Ours}
            & \textbf{13.4} & \textbf{37.3}
            & \textbf{21.0} & \textbf{25.6}
            & \textbf{25.5} & \textbf{26.2} & \textbf{35.8} \\
            \bottomrule
        \end{tabular}
        }

        \label{tab:comp_layerdiffuse}
        \vspace{-8pt}
    \end{minipage}
    \hfill
    \begin{minipage}{0.47\textwidth}
        \centering

        \captionof{table}{
        \textbf{Comparison with Qwen~\cite{wu2025qwen} for layer editing.}
        R: recolor; M: movement; C: joint recolor+movement.
        }
        \vspace{-6pt}

        \resizebox{\textwidth}{!}{
        \begin{tabular}{lccc}
            \toprule
            \textbf{Method} & \textbf{R} & \textbf{M} & \textbf{C} \\
            & CLIP-FID/FID & CLIP-FID/FID & CLIP-FID/FID \\
            \midrule
            Instr. (Qwen) & 13.2/102.9 & 13.4/101.4 & 15.8/110.8 \\
            Casc. Layer (Qwen)  & 9.5/88.8   & 8.5/83.6    & 8.8/86.9 \\
            \textbf{Layer+VE (Ours)}
            & \textbf{8.3}/\textbf{71.7}
            & \textbf{6.5}/\textbf{68.4}
            & \textbf{6.4}/\textbf{71.1} \\
            \bottomrule
        \end{tabular}
        }

        \label{tab:quant:editing}
    \end{minipage}

\end{table*}

\subsection{Creative Multi-object Editing Applications}
\vspace{-8pt}

\begin{figure*}[t]
  \centering
    \includegraphics[width=1\linewidth]{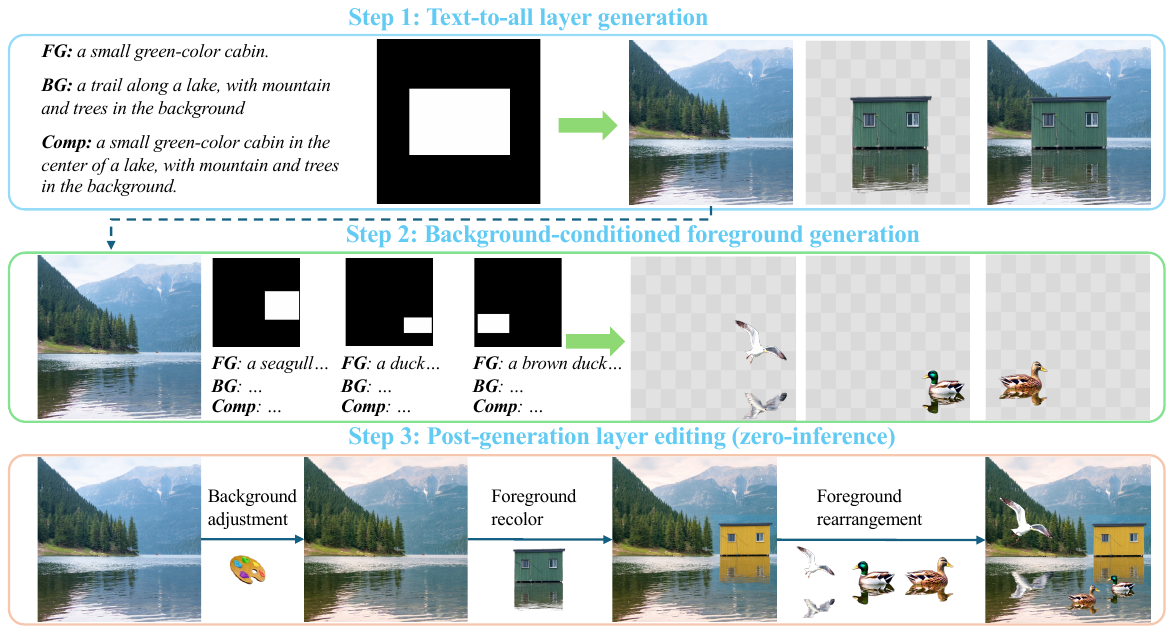}
   \vspace{-15pt}
   \caption{\textbf{Diverse multi-object creative applications driven by our model.} We leverage both Text2All and FG\_Gen modes to jointly guide the synthesis process, unlocking a broader range of editing possibilities and producing diverse, visually appealing results.}
   \label{fig:create_app}
   \vspace{-12pt}
\end{figure*}

Our framework supports generating multiple layers with realistic visual effects under three different generation modes, enabling natural post-generation layer editing, as shown in \cref{fig:teaser_image} and \cref{fig:com_others}. In this section, we further demonstrate that, benefiting from the highly consistent and visually coherent results across different modes, our framework allows flexible cross-mode collaborative editing, unlocking a broader range of creative possibilities. As illustrated in \cref{fig:create_app}, combining the Text2All and FG\_Gen modes yields diverse and harmonized editing outcomes.

\section{Conclusion and Discussion}
\label{sec:conclusion}
\vspace{-10pt}

In this work, we present \ours{}, a novel and unified framework for controllable layered image generation and editing. To facilitate research in this direction, we introduce the \dataset{} dataset and the \benchmark{} evaluation benchmark. A core advantage of our approach is the explicit binding of high-fidelity, context-aware visual effects to the FG layer. This design is highly practical for real-world applications, addressing the prevalent need for single-object composition and local editing—such as translation and resizing—without requiring model re-inference. Within this common ``FG over BG'' setting, \ours{} significantly outperforms existing methods in preserving object identity and maintaining perceptual realism. Extending our framework to dynamically adapt visual effects across complex multi-object interactions and intricate scene geometries remains an exciting direction for future work.

\bibliographystyle{splncs04}
\bibliography{main}

@String(CVPR= {IEEE Conf. Comput. Vis. Pattern Recog.})

@String(ECCV= {Eur. Conf. Comput. Vis.})

@String(CVPR  = {CVPR})

@String(ECCV  = {ECCV})

@article{su2024roformer,
  title={Roformer: Enhanced transformer with rotary position embedding},
  author={Su, Jianlin and Ahmed, Murtadha and Lu, Yu and Pan, Shengfeng and Bo, Wen and Liu, Yunfeng},
  journal={Neurocomputing},
  volume={568},
  pages={127063},
  year={2024},
  publisher={Elsevier}
}

@inproceedings{pu2025art,
  title={Art: Anonymous region transformer for variable multi-layer transparent image generation},
  author={Pu, Yifan and Zhao, Yiming and Tang, Zhicong and Yin, Ruihong and Ye, Haoxing and Yuan, Yuhui and Chen, Dong and Bao, Jianmin and Zhang, Sirui and Wang, Yanbin and others},
  booktitle={Proceedings of the Computer Vision and Pattern Recognition Conference},
  pages={7952--7962},
  year={2025}
}

@inproceedings{huang2025dreamlayer,
  title={DreamLayer: Simultaneous Multi-Layer Generation via Diffusion Model},
  author={Huang, Junjia and Yan, Pengxiang and Cai, Jinhang and Liu, Jiyang and Wang, Zhao and Wang, Yitong and Wu, Xinglong and Li, Guanbin},
  booktitle={Proceedings of the IEEE/CVF International Conference on Computer Vision},
  pages={3357--3366},
  year={2025}
}

@article{yin2025qwen,
  title={Qwen-Image-Layered: Towards Inherent Editability via Layer Decomposition},
  author={Yin, Shengming and Zhang, Zekai and Tang, Zecheng and Gao, Kaiyuan and Xu, Xiao and Yan, Kun and Li, Jiahao and Chen, Yilei and Chen, Yuxiang and Shum, Heung-Yeung and others},
  journal={arXiv preprint arXiv:2512.15603},
  year={2025}
}

@inproceedings{rombach2022high,
  title={High-resolution image synthesis with latent diffusion models},
  author={Rombach, Robin and Blattmann, Andreas and Lorenz, Dominik and Esser, Patrick and Ommer, Bj{\"o}rn},
  booktitle={Proceedings of the IEEE/CVF conference on computer vision and pattern recognition},
  pages={10684--10695},
  year={2022}
}

@article{podell2023sdxl,
  title={Sdxl: Improving latent diffusion models for high-resolution image synthesis},
  author={Podell, Dustin and English, Zion and Lacey, Kyle and Blattmann, Andreas and Dockhorn, Tim and M{\"u}ller, Jonas and Penna, Joe and Rombach, Robin},
  journal={arXiv preprint arXiv:2307.01952},
  year={2023}
}

@inproceedings{esser2024scaling,
  title={Scaling rectified flow transformers for high-resolution image synthesis},
  author={Esser, Patrick and Kulal, Sumith and Blattmann, Andreas and Entezari, Rahim and M{\"u}ller, Jonas and Saini, Harry and Levi, Yam and Lorenz, Dominik and Sauer, Axel and Boesel, Frederic and others},
  booktitle={Forty-first international conference on machine learning},
  year={2024}
}

@inproceedings{ramesh2021zero,
  title={Zero-shot text-to-image generation},
  author={Ramesh, Aditya and Pavlov, Mikhail and Goh, Gabriel and Gray, Scott and Voss, Chelsea and Radford, Alec and Chen, Mark and Sutskever, Ilya},
  booktitle={International conference on machine learning},
  pages={8821--8831},
  year={2021},
  organization={Pmlr}
}

@article{wu2025qwen,
  title={Qwen-image technical report},
  author={Wu, Chenfei and Li, Jiahao and Zhou, Jingren and Lin, Junyang and Gao, Kaiyuan and Yan, Kun and Yin, Sheng-ming and Bai, Shuai and Xu, Xiao and Chen, Yilei and others},
  journal={arXiv preprint arXiv:2508.02324},
  year={2025}
}

@misc{openai_gptimage1_2025,
  title        = {GPT Image 1},
  author       = {{OpenAI}},
  year         = {2025},
  howpublished = {\url{https://platform.openai.com/docs/models/gpt-image-1}},
  note         = {Accessed: 2025-11-13}
}

@article{labs2025flux,
  title={FLUX. 1 Kontext: Flow Matching for In-Context Image Generation and Editing in Latent Space},
  author={Labs, Black Forest and Batifol, Stephen and Blattmann, Andreas and Boesel, Frederic and Consul, Saksham and Diagne, Cyril and Dockhorn, Tim and English, Jack and English, Zion and Esser, Patrick and others},
  journal={arXiv preprint arXiv:2506.15742},
  year={2025}
}

@misc{huggingface2025flux,
  title        = {Flux Family Models},
  author       = {{Hugging Face}},
  howpublished = {\url{https://huggingface.co/docs/diffusers/main/en/api/pipelines/flux}},
  year         = {2025}
}

@article{ye2025imgedit,
  title={Imgedit: A unified image editing dataset and benchmark},
  author={Ye, Yang and He, Xianyi and Li, Zongjian and Lin, Bin and Yuan, Shenghai and Yan, Zhiyuan and Hou, Bohan and Yuan, Li},
  journal={arXiv preprint arXiv:2505.20275},
  year={2025}
}

@article{zhao2024ultraedit,
  title={Ultraedit: Instruction-based fine-grained image editing at scale},
  author={Zhao, Haozhe and Ma, Xiaojian Shawn and Chen, Liang and Si, Shuzheng and Wu, Rujie and An, Kaikai and Yu, Peiyu and Zhang, Minjia and Li, Qing and Chang, Baobao},
  journal={Advances in Neural Information Processing Systems},
  volume={37},
  pages={3058--3093},
  year={2024}
}

@article{liu2025step1x,
  title={Step1x-edit: A practical framework for general image editing},
  author={Liu, Shiyu and Han, Yucheng and Xing, Peng and Yin, Fukun and Wang, Rui and Cheng, Wei and Liao, Jiaqi and Wang, Yingming and Fu, Honghao and Han, Chunrui and others},
  journal={arXiv preprint arXiv:2504.17761},
  year={2025}
}

@article{ravi2024sam,
  title={Sam 2: Segment anything in images and videos},
  author={Ravi, Nikhila and Gabeur, Valentin and Hu, Yuan-Ting and Hu, Ronghang and Ryali, Chaitanya and Ma, Tengyu and Khedr, Haitham and R{\"a}dle, Roman and Rolland, Chloe and Gustafson, Laura and others},
  journal={arXiv preprint arXiv:2408.00714},
  year={2024}
}

@article{ren2024grounded,
  title={Grounded sam: Assembling open-world models for diverse visual tasks},
  author={Ren, Tianhe and Liu, Shilong and Zeng, Ailing and Lin, Jing and Li, Kunchang and Cao, He and Chen, Jiayu and Huang, Xinyu and Chen, Yukang and Yan, Feng and others},
  journal={arXiv preprint arXiv:2401.14159},
  year={2024}
}

@article{wei2025omnieraser,
  title={OmniEraser: Remove Objects and Their Effects in Images with Paired Video-Frame Data},
  author={Wei, Runpu and Yin, Zijin and Zhang, Shuo and Zhou, Lanxiang and Wang, Xueyi and Ban, Chao and Cao, Tianwei and Sun, Hao and He, Zhongjiang and Liang, Kongming and others},
  journal={arXiv preprint arXiv:2501.07397},
  year={2025}
}

@article{zhao2025objectclear,
  title={ObjectClear: Complete Object Removal via Object-Effect Attention},
  author={Zhao, Jixin and Zhou, Shangchen and Wang, Zhouxia and Yang, Peiqing and Loy, Chen Change},
  journal={arXiv preprint arXiv:2505.22636},
  year={2025}
}

@inproceedings{tudosiu2024mulan,
  title={Mulan: A multi layer annotated dataset for controllable text-to-image generation},
  author={Tudosiu, Petru-Daniel and Yang, Yongxin and Zhang, Shifeng and Chen, Fei and McDonagh, Steven and Lampouras, Gerasimos and Iacobacci, Ignacio and Parisot, Sarah},
  booktitle={Proceedings of the IEEE/CVF Conference on Computer Vision and Pattern Recognition},
  pages={22413--22422},
  year={2024}
}

@article{zhang2023text2layer,
  title={Text2layer: Layered image generation using latent diffusion model},
  author={Zhang, Xinyang and Zhao, Wentian and Lu, Xin and Chien, Jeff},
  journal={arXiv preprint arXiv:2307.09781},
  year={2023}
}

@article{zhong2025tp,
  title={TP-Blend: Textual-Prompt Attention Pairing for Precise Object-Style Blending in Diffusion Models},
  author={Zhong, Yichuan and Tian, Yapeng and others},
  journal={Transactions on Machine Learning Research},
  year={2025}
}

@article{zhang2024transparent,
  title={Transparent image layer diffusion using latent transparency},
  author={Zhang, Lvmin and Agrawala, Maneesh},
  journal={arXiv preprint arXiv:2402.17113},
  year={2024}
}

@article{dalva2024layerfusion,
  title={Layerfusion: Harmonized multi-layer text-to-image generation with generative priors},
  author={Dalva, Yusuf and Li, Yijun and Liu, Qing and Zhao, Nanxuan and Zhang, Jianming and Lin, Zhe and Yanardag, Pinar},
  journal={arXiv preprint arXiv:2412.04460},
  year={2024}
}

@article{fontanella2024generating,
  title={Generating compositional scenes via Text-to-image RGBA Instance Generation},
  author={Fontanella, Alessandro and Tudosiu, Petru-Daniel and Yang, Yongxin and Zhang, Shifeng and Parisot, Sarah},
  journal={Advances in Neural Information Processing Systems},
  volume={37},
  pages={43864--43893},
  year={2024}
}

@article{huang2025psdiffusion,
  title={PSDiffusion: Harmonized Multi-Layer Image Generation via Layout and Appearance Alignment},
  author={Huang, Dingbang and Li, Wenbo and Zhao, Yifei and Pan, Xinyu and Zeng, Yanhong and Dai, Bo},
  journal={arXiv preprint arXiv:2505.11468},
  year={2025}
}

@inproceedings{chen2025unireal,
  title={Unireal: Universal image generation and editing via learning real-world dynamics},
  author={Chen, Xi and Zhang, Zhifei and Zhang, He and Zhou, Yuqian and Kim, Soo Ye and Liu, Qing and Li, Yijun and Zhang, Jianming and Zhao, Nanxuan and Wang, Yilin and others},
  booktitle={Proceedings of the Computer Vision and Pattern Recognition Conference},
  pages={12501--12511},
  year={2025}
}

@inproceedings{yang2025generative,
  title={Generative Image Layer Decomposition with Visual Effects},
  author={Yang, Jinrui and Liu, Qing and Li, Yijun and Kim, Soo Ye and Pakhomov, Daniil and Ren, Mengwei and Zhang, Jianming and Lin, Zhe and Xie, Cihang and Zhou, Yuyin},
  booktitle={Proceedings of the Computer Vision and Pattern Recognition Conference},
  pages={7643--7653},
  year={2025}
}

@article{kang2025layeringdiff,
  title={LayeringDiff: Layered Image Synthesis via Generation, then Disassembly with Generative Knowledge},
  author={Kang, Kyoungkook and Sim, Gyujin and Kim, Geonung and Kim, Donguk and Nam, Seungho and Cho, Sunghyun},
  journal={arXiv preprint arXiv:2501.01197},
  year={2025}
}

@inproceedings{kirillov2023segment,
  title={Segment anything},
  author={Kirillov, Alexander and Mintun, Eric and Ravi, Nikhila and Mao, Hanzi and Rolland, Chloe and Gustafson, Laura and Xiao, Tete and Whitehead, Spencer and Berg, Alexander C and Lo, Wan-Yen and others},
  booktitle={Proceedings of the IEEE/CVF international conference on computer vision},
  pages={4015--4026},
  year={2023}
}

@inproceedings{zhang2023adding,
  title={Adding conditional control to text-to-image diffusion models},
  author={Zhang, Lvmin and Rao, Anyi and Agrawala, Maneesh},
  booktitle={Proceedings of the IEEE/CVF international conference on computer vision},
  pages={3836--3847},
  year={2023}
}

@inproceedings{zhuang2024task,
  title={A task is worth one word: Learning with task prompts for high-quality versatile image inpainting},
  author={Zhuang, Junhao and Zeng, Yanhong and Liu, Wenran and Yuan, Chun and Chen, Kai},
  booktitle={European Conference on Computer Vision},
  pages={195--211},
  year={2024},
  organization={Springer}
}

@inproceedings{ju2024brushnet,
  title={Brushnet: A plug-and-play image inpainting model with decomposed dual-branch diffusion},
  author={Ju, Xuan and Liu, Xian and Wang, Xintao and Bian, Yuxuan and Shan, Ying and Xu, Qiang},
  booktitle={European Conference on Computer Vision},
  pages={150--168},
  year={2024},
  organization={Springer}
}

@inproceedings{lin2014microsoft,
  title={Microsoft coco: Common objects in context},
  author={Lin, Tsung-Yi and Maire, Michael and Belongie, Serge and Hays, James and Perona, Pietro and Ramanan, Deva and Doll{\'a}r, Piotr and Zitnick, C Lawrence},
  booktitle={European conference on computer vision},
  pages={740--755},
  year={2014},
  organization={Springer}
}

@article{liu2025seg,
  title={Seg-zero: Reasoning-chain guided segmentation via cognitive reinforcement},
  author={Liu, Yuqi and Peng, Bohao and Zhong, Zhisheng and Yue, Zihao and Lu, Fanbin and Yu, Bei and Jia, Jiaya},
  journal={arXiv preprint arXiv:2503.06520},
  year={2025}
}

@article{wang2022instance,
  title={Instance shadow detection with a single-stage detector},
  author={Wang, Tianyu and Hu, Xiaowei and Heng, Pheng-Ann and Fu, Chi-Wing},
  journal={IEEE transactions on pattern analysis and machine intelligence},
  volume={45},
  number={3},
  pages={3259--3273},
  year={2022},
  publisher={IEEE}
}

@article{ke2025marigold,
  title={Marigold: Affordable Adaptation of Diffusion-Based Image Generators for Image Analysis},
  author={Ke, Bingxin and Qu, Kevin and Wang, Tianfu and Metzger, Nando and Huang, Shengyu and Li, Bo and Obukhov, Anton and Schindler, Konrad},
  journal={arXiv preprint arXiv:2505.09358},
  year={2025}
}

@inproceedings{gao2024multi,
  title={Multi-scale and detail-enhanced segment anything model for salient object detection},
  author={Gao, Shixuan and Zhang, Pingping and Yan, Tianyu and Lu, Huchuan},
  booktitle={Proceedings of the 32nd ACM International Conference on Multimedia},
  pages={9894--9903},
  year={2024}
}

@article{chen2024expanding,
  title={Expanding performance boundaries of open-source multimodal models with model, data, and test-time scaling},
  author={Chen, Zhe and Wang, Weiyun and Cao, Yue and Liu, Yangzhou and Gao, Zhangwei and Cui, Erfei and Zhu, Jinguo and Ye, Shenglong and Tian, Hao and Liu, Zhaoyang and others},
  journal={arXiv preprint arXiv:2412.05271},
  year={2024}
}

@article{bai2025qwen2,
  title={Qwen2. 5-vl technical report},
  author={Bai, Shuai and Chen, Keqin and Liu, Xuejing and Wang, Jialin and Ge, Wenbin and Song, Sibo and Dang, Kai and Wang, Peng and Wang, Shijie and Tang, Jun and others},
  journal={arXiv preprint arXiv:2502.13923},
  year={2025}
}

@misc{unsplash,
  author       = "{Unsplash}",
  title        = "Unsplash: Free High-Resolution Photos",
  howpublished = "\url{https://unsplash.com/}"
}

@inproceedings{winter2024objectdrop,
  title={Objectdrop: Bootstrapping counterfactuals for photorealistic object removal and insertion},
  author={Winter, Daniel and Cohen, Matan and Fruchter, Shlomi and Pritch, Yael and Rav-Acha, Alex and Hoshen, Yedid},
  booktitle={European Conference on Computer Vision},
  pages={112--129},
  year={2024},
  organization={Springer}
}

@article{heusel2017gans,
  title={Gans trained by a two time-scale update rule converge to a local nash equilibrium},
  author={Heusel, Martin and Ramsauer, Hubert and Unterthiner, Thomas and Nessler, Bernhard and Hochreiter, Sepp},
  journal={Advances in neural information processing systems},
  volume={30},
  year={2017}
}

@inproceedings{parmar2021cleanfid,
  title={On Aliased Resizing and Surprising Subtleties in GAN Evaluation},
  author={Parmar, Gaurav and Zhang, Richard and Zhu, Jun-Yan},
  booktitle={CVPR},
  year={2022}
}

@article{yang2025complexedit,
  title={Complexedit: Cot-like instruction generation for complexity-controllable image editing benchmark},
  author={Yang, Siwei and Hui, Mude and Zhao, Bingchen and Zhou, Yuyin and Ruiz, Nataniel and Xie, Cihang},
  journal={arXiv preprint arXiv:2504.13143},
  year={2025}
}

@inproceedings{zhang2018unreasonable,
  title={The unreasonable effectiveness of deep features as a perceptual metric},
  author={Zhang, Richard and Isola, Phillip and Efros, Alexei A and Shechtman, Eli and Wang, Oliver},
  booktitle={Proceedings of the IEEE conference on computer vision and pattern recognition},
  pages={586--595},
  year={2018}
}

@inproceedings{peebles2023scalable,
  title={Scalable diffusion models with transformers},
  author={Peebles, William and Xie, Saining},
  booktitle={Proceedings of the IEEE/CVF International Conference on Computer Vision},
  pages={4195--4205},
  year={2023}
}

@article{lipman2022flow,
  title={Flow matching for generative modeling},
  author={Lipman, Yaron and Chen, Ricky TQ and Ben-Hamu, Heli and Nickel, Maximilian and Le, Matt},
  journal={arXiv preprint arXiv:2210.02747},
  year={2022}
}

@article{raffel2020exploring,
  title={Exploring the limits of transfer learning with a unified text-to-text transformer},
  author={Raffel, Colin and Shazeer, Noam and Roberts, Adam and Lee, Katherine and Narang, Sharan and Matena, Michael and Zhou, Yanqi and Li, Wei and Liu, Peter J},
  journal={Journal of machine learning research},
  volume={21},
  number={140},
  pages={1--67},
  year={2020}
}

@article{zhang2025context,
  title={In-context edit: Enabling instructional image editing with in-context generation in large scale diffusion transformer},
  author={Zhang, Zechuan and Xie, Ji and Lu, Yu and Yang, Zongxin and Yang, Yi},
  journal={arXiv preprint arXiv:2504.20690},
  year={2025}
}

@inproceedings{yu2025anyedit,
  title={Anyedit: Mastering unified high-quality image editing for any idea},
  author={Yu, Qifan and Chow, Wei and Yue, Zhongqi and Pan, Kaihang and Wu, Yang and Wan, Xiaoyang and Li, Juncheng and Tang, Siliang and Zhang, Hanwang and Zhuang, Yueting},
  booktitle={Proceedings of the Computer Vision and Pattern Recognition Conference},
  pages={26125--26135},
  year={2025}
}

@inproceedings{brooks2023instructpix2pix,
  title={Instructpix2pix: Learning to follow image editing instructions},
  author={Brooks, Tim and Holynski, Aleksander and Efros, Alexei A},
  booktitle={Proceedings of the IEEE/CVF conference on computer vision and pattern recognition},
  pages={18392--18402},
  year={2023}
}

@article{lin2025uniworld,
  title={Uniworld: High-resolution semantic encoders for unified visual understanding and generation},
  author={Lin, Bin and Li, Zongjian and Cheng, Xinhua and Niu, Yuwei and Ye, Yang and He, Xianyi and Yuan, Shenghai and Yu, Wangbo and Wang, Shaodong and Ge, Yunyang and others},
  journal={arXiv preprint arXiv:2506.03147},
  year={2025}
}

@article{wu2025omnigen2,
  title={OmniGen2: Exploration to Advanced Multimodal Generation},
  author={Wu, Chenyuan and Zheng, Pengfei and Yan, Ruiran and Xiao, Shitao and Luo, Xin and Wang, Yueze and Li, Wanli and Jiang, Xiyan and Liu, Yexin and Zhou, Junjie and others},
  journal={arXiv preprint arXiv:2506.18871},
  year={2025}
}

@article{deng2025emerging,
  title={Emerging properties in unified multimodal pretraining},
  author={Deng, Chaorui and Zhu, Deyao and Li, Kunchang and Gou, Chenhui and Li, Feng and Wang, Zeyu and Zhong, Shu and Yu, Weihao and Nie, Xiaonan and Song, Ziang and others},
  journal={arXiv preprint arXiv:2505.14683},
  year={2025}
}

@article{ghosh2023geneval,
  title={Geneval: An object-focused framework for evaluating text-to-image alignment},
  author={Ghosh, Dhruba and Hajishirzi, Hannaneh and Schmidt, Ludwig},
  journal={Advances in Neural Information Processing Systems},
  volume={36},
  pages={52132--52152},
  year={2023}
}

@inproceedings{chen2024pixart,
  title={Pixart-sigma: Weak-to-strong training of diffusion transformer for 4k text-to-image generation},
  author={Chen, Junsong and Ge, Chongjian and Xie, Enze and Wu, Yue and Yao, Lewei and Ren, Xiaozhe and Wang, Zhongdao and Luo, Ping and Lu, Huchuan and Li, Zhenguo},
  booktitle={ECCV},
  year={2024}
}

@article{emu3,
  title={Emu3: Next-token prediction is all you need},
  author={Wang, Xinlong and Zhang, Xiaosong and Luo, Zhengxiong and Sun, Quan and Cui, Yufeng and Wang, Jinsheng and Zhang, Fan and Wang, Yueze and Li, Zhen and Yu, Qiying and others},
  journal={arXiv preprint arxiv:2409.18869},
  year={2024}
}

@misc{flux,
  title = {FLUX},
  author = {Black Forest Labs},
  year = {2024},
  url = {https://github.com/black-forest-labs/flux}
}

@article{seed-x,
  title={Seed-x: Multimodal models with unified multi-granularity comprehension and generation},
  author={Ge, Yuying and Zhao, Sijie and Zhu, Jinguo and Ge, Yixiao and Yi, Kun and Song, Lin and Li, Chen and Ding, Xiaohan and Shan, Ying},
  journal={arXiv preprint arxiv:2404.14396},
  year={2024}
}

@article{transfusion,
  title={Transfusion: Predict the next token and diffuse images with one multi-modal model},
  author={Zhou, Chunting and Yu, Lili and Babu, Arun and Tirumala, Kushal and Yasunaga, Michihiro and Shamis, Leonid and Kahn, Jacob and Ma, Xuezhe and Zettlemoyer, Luke and Levy, Omer},
  journal={arXiv preprint arxiv:2408.11039},
  year={2024}
}

@article{show-o,
  title={Show-o: One single transformer to unify multimodal understanding and generation},
  author={Xie, Jinheng and Mao, Weijia and Bai, Zechen and Zhang, David Junhao and Wang, Weihao and Lin, Kevin Qinghong and Gu, Yuchao and Chen, Zhijie and Yang, Zhenheng and Shou, Mike Zheng},
  journal={arXiv preprint arxiv:2408.12528},
  year={2024}
}

@article{januspro2025,
  title={Janus-Pro: Unified Multimodal Understanding and Generation with Data and Model Scaling},
  author={Xiaokang Chen and Chengyue Wu and Zhiyu Wu and Yiyang Ma and Xingchao Liu and Zizheng Pan and Wen Liu and Zhenda Xie and Xingkai Yu and Chong Ruan and Ping Luo},
  journal={arXiv preprint arXiv:2501.17811},
  year={2025},
}

@article{pan2025transfer,
  title={Transfer between Modalities with MetaQueries},
  author={Pan, Xichen and Shukla, Satya Narayan and Singh, Aashu and Zhao, Zhuokai and Mishra, Shlok Kumar and Wang, Jialiang and Xu, Zhiyang and Chen, Jiuhai and Li, Kunpeng and Juefei-Xu, Felix and Hou, Ji and Xie, Saining},
  year={2025},
  journal={arXiv preprint arXiv:2504.06256}
}

@article{dalle3,
  title={Improving image generation with better captions},
  author={Betker, James and Goh, Gabriel and Jing, Li and Brooks, Tim and Wang, Jianfeng and Li, Linjie and Ouyang, Long and Zhuang, Juntang and Lee, Joyce and Guo, Yufei and others},
  journal={OpenAI blog},
  year={2023}
}

@inproceedings{SD3,
  title={Scaling rectified flow transformers for high-resolution image synthesis},
  author={Esser, Patrick and Kulal, Sumith and Blattmann, Andreas and Entezari, Rahim and M{\"u}ller, Jonas and Saini, Harry and Levi, Yam and Lorenz, Dominik and Sauer, Axel and Boesel, Frederic and others},
  booktitle={ICML},
  year={2024}
}

\appendix
\renewcommand{\thesection}{\arabic{section}}
\setcounter{section}{5}

\clearpage
\begin{center}
  {\LARGE\bfseries Appendix}\\[0.5em]
\end{center}
\vspace{1.2em}

\section{Video Demo}

We provide a demonstration video, \emph{Video Demo}, in the Supplementary Material, which offers an intuitive visualization of how our 
three generation modes operate in practice.

\section{\ours{} Framework: Detailed Architecture}
\label{app:framework_details}

We provide additional details on how heterogeneous image and text inputs are
tokenized, tagged with our task-aware embeddings, and assembled into the
single sequence consumed by the DiT backbone. Throughout this section we
follow the notation introduced in~\cref{tab:modes}: an image $\mathbf{u}\in
\{\mathbf{x}_t^{\text{comp}}, \mathbf{x}_t^{\text{bg}},
\mathbf{x}_t^{\text{fg+ve}}, \mathbf{c}_\text{bg}, \mathbf{c}_\text{fg},
\mathbf{c}_\text{mask}\}$ is referred to as a \emph{frame}, and the model
processes a variable-length set of $F$ such frames per sample.

\subsection{Image Tokenization}
\label{app:tokenization}

All image inputs (both conditional and noisy targets) live in a shared latent
space provided by a frozen RGBA VAE encoder $\mathcal{E}_{\text{VAE}}$. Given
a raw frame $\mathbf{u}$, we obtain
\begin{equation}
    \mathbf{z} \;=\; \mathcal{E}_{\text{VAE}}(\mathbf{u})
    \;\in\;\mathbb{R}^{C\times H\times W},
\end{equation}
including the mask condition $\mathbf{c}_\text{mask}$, which we encode with
the same VAE so that all conditioning lives in the same latent manifold.
Each frame latent is then patchified and linearly projected to the model
dimension $D$ by an embedder $\mathbf{W}_{\!x}$, yielding $N_{\text{img}}$
tokens per frame. Stacking the $F$ frames gives the image token block
$\mathbf{H}_{\text{img}}\in\mathbb{R}^{F N_{\text{img}}\times D}$.

\subsection{Four Embeddings: Type, IO, Position, Timestep}
\label{app:embeddings}

Because all frames---inputs and noisy targets, BG/FG/composite/mask---are
flattened into a single token stream, the network must be told (i) what each
frame \emph{is}, (ii) whether it is being conditioned on or denoised, and
(iii) where each token sits in space and in the diffusion process. We
therefore augment every image token belonging to frame $f\in\{1,\dots,F\}$
with four embeddings.

\paragraph{Type Embedding $\mathbf{e}^{\text{type}}$.}
This is the central mechanism that lets a single network reason jointly over
the BG, FG, composite, and mask layers. We assign each frame a discrete type
$\tau_f \in \mathcal{T}=\{\textsc{Asset},\,\textsc{Canvas}_{\text{bg}},\,
\textsc{Canvas}_{\text{comp}},\,\textsc{Canvas}_{\text{fg{+}ve}},\,
\textsc{Control}\}$,
where \textsc{Asset} marks a clean FG reference, the three
\textsc{Canvas} variants enumerate the BG, composite, and FG-with-visual-effect
canvases, and \textsc{Control} marks the mask frame. A learned embedding
$\phi_{\text{type}}\colon\mathcal{T}\to\mathbb{R}^{D}$ is then broadcast over
all tokens of frame $f$:
\begin{equation}
    \mathbf{e}^{\text{type}}_{f,i} \;=\; \phi_{\text{type}}(\tau_f).
\end{equation}
Without this embedding, the transformer cannot tell apart frames of
different semantic roles---\eg, the target composite
$\mathbf{x}_t^{\text{comp}}$ vs.\ the conditional background
$\mathbf{c}_\text{bg}$.

\paragraph{IO Embedding $\mathbf{e}^{\text{io}}$.}
This embedding tells the model whether a frame is an \emph{input}
($\mathbf{c}\in\mathbf{C}$) to be preserved or an \emph{output}
($\mathbf{x}\in\mathbf{X}_t$) to be denoised. We assign each frame an
indicator $\rho_f \in \{\textsc{Input}, \textsc{Output}\}$, and a learned
$\phi_{\text{io}}$ produces
\begin{equation}
    \mathbf{e}^{\text{io}}_{f,i} \;=\; \phi_{\text{io}}(\rho_f).
\end{equation}
Together with $\mathbf{e}^{\text{type}}$, this single scalar flag is the only difference between the three modes
in~\cref{tab:modes}: under \textsc{FG\_Gen}, the BG frame carries
$\rho{=}\textsc{Input}$; under \textsc{BG\_Gen}, the FG frame does; and
under \textsc{Text2All}, all image frames carry $\rho{=}\textsc{Output}$. By varying the assignment
$\{(\tau_f,\rho_f)\}_{f=1}^{F}$, the same set of weights learns to support
every editing mode without any architectural change.

\paragraph{Spatial--Temporal Position Embedding $\mathbf{e}^{\text{pos}}$.}
A key technical ingredient of \ours{} is a unified \emph{3D rotary position
embedding} (RoPE)~\cite{su2024roformer} over the joint
$(\text{height},\text{width},\text{frame})$ axes. Concretely, for a token at
spatial location $(h,w)$ within frame index $f$, we apply RoPE on the
$\mathrm{Q},\mathrm{K}$ projections of every attention head:
\begin{equation}
    \mathbf{e}^{\text{pos}}_{f,h,w}
    \;=\; \mathrm{RoPE}_{\text{3D}}\bigl(h,\,w,\,\eta_f\bigr),
\end{equation}
where $\eta_f$ is a per-frame index. Adding the frame index as an explicit
positional axis lets the model preserve the 2D spatial structure within
each frame while still knowing which layer (BG, FG, composite, mask) each
token belongs to, so cross-frame correspondences can be consistently
modeled in attention.

\paragraph{Timestep Embedding $\mathbf{e}^{\text{time}}$.}
The diffusion step $t$ is encoded by a sinusoidal positional encoding
followed by an MLP, $\mathbf{e}^{\text{time}}(t) \in \mathbb{R}^{D}$, and
broadcast over all visual tokens.

\medskip
The Type, IO, and Timestep embeddings are aggregated into a per-token
modulation
\begin{equation}
    \mathbf{m}_{f,i}
    \;=\; \mathrm{Norm}\!\bigl(
    \mathbf{e}^{\text{time}}(t)
    + \mathbf{e}^{\text{type}}_{f,i}
    + \mathbf{e}^{\text{io}}_{f,i}\bigr),
    \label{eq:modulation}
\end{equation}
which is fed to every DiT block to modulate its computation, while
$\mathbf{e}^{\text{pos}}$ is injected inside attention via 3D RoPE.

\subsection{Sequence Assembly and DiT Forward Pass}
\label{app:sequence}

Let $\mathbf{H}_\text{cond}\in\mathbb{R}^{L_\text{cond}\times D}$ denote
the condition tokens encoded from the prompt $\mathbf{c}_\text{txt}$. We
concatenate them with the frame-wise image tokens into a single sequence
\begin{equation}
    \mathbf{S}
    \;=\; \bigl[\,\mathbf{H}_\text{cond};\;
                 \mathbf{H}^{(1)}_\text{img};\;\dots;\;
                 \mathbf{H}^{(F)}_\text{img}\,\bigr],
\end{equation}
which is processed by a stack of DiT blocks with full bidirectional
self-attention---so that every output frame jointly attends to all
conditional frames and tokens. The output tokens for frames flagged $\rho_f{=}\textsc{Output}$ are then
mapped to the noise prediction $\boldsymbol{\epsilon}_\theta$ used in the flow-matching loss.

\paragraph{Why Type and IO Embeddings Matter.}
The Type and IO embeddings together act as a soft programmatic interface to
the network: $\phi_{\text{type}}$ specifies the \emph{semantic role} of each
frame (FG asset, BG canvas, composite canvas, mask), while $\phi_{\text{io}}$
specifies its \emph{causal role} (given vs.\ generated). At inference time,
switching between \textsc{FG\_Gen}, \textsc{BG\_Gen}, and \textsc{Text2All}
requires \emph{no} parameter update---only a different assignment of
$\{(\tau_f,\rho_f)\}_{f=1}^{F}$ over the input frames. This decoupling is
what enables \ours{} to address real-world editing needs with
a single model.

\section{More Experiments}

\subsection{Comparison on Public Benchmarks}

While primarily designed for the novel task of layer generation, we further investigate its versatility by adapting it to standard public benchmarks: ImgEdit-Bench~\cite{ye2025imgedit} and GenEval~\cite{ghosh2023geneval}, aligning our generation modes (FG\_Gen, BG\_Gen, and Text2All) with their corresponding benchmark tasks (``Addition'', ``Background'', and standard text-to-image generation).

As shown in \cref{tab:add_background} and \cref{tab:geneval}, our model demonstrates strong performance on the composite images. It is important to emphasize that current SOTA methods benefit from significantly larger training corpora and model sizes, and are aligned with these standard benchmark protocols. In contrast, \ours{} approaches these tasks without specific optimization, yet still maintains highly competitive results. Moreover, a key advantage of our approach is its ability to perform layer generation with visual effects, a capability not supported by existing models. The effectiveness of layer representations---and their clear benefits for subsequent editing quality---has been thoroughly validated in the above, highlighting a unique strength of our method.

\begin{table*}[ht]
\centering
\caption{\textbf{Evaluation of image editing ability on ImgEdit-Bench~\cite{ye2025imgedit}.} ``Addition'' corresponds to FG\_Gen, and ``Background'' corresponds to BG\_Gen.}
\resizebox{0.8\columnwidth}{!}{
\begin{tabular}{lcc|lcc}
\toprule
\textbf{Model} & \textbf{Addition} & \textbf{Background} & \textbf{Model} & \textbf{Addition} & \textbf{Background} \\
\midrule
Instruct-P2P~\cite{brooks2023instructpix2pix} & 2.45 & 1.44 & UniWorld-V1~\cite{lin2025uniworld}  & 3.82 & 2.99 \\
AnyEdit~\cite{yu2025anyedit}                  & 3.18 & 2.24 & BAGEL~\cite{deng2025emerging}       & 3.81 & 3.39 \\
UltraEdit~\cite{zhao2024ultraedit}            & 3.44 & 2.83 & OmniGen2~\cite{wu2025omnigen2}      & 3.57 & 3.57 \\
ICEdit~\cite{zhang2025context}                & 3.58 & 3.08 & Kontext-dev~\cite{labs2025flux}     & 3.83 & 3.98 \\
Step1X-Edit~\cite{liu2025step1x}              & 3.88 & 3.16 & \ours{}                             & 3.86 & 3.32 \\
\bottomrule
\end{tabular}
}
\label{tab:add_background}
\end{table*}

\begin{table*}[ht]
    \centering
    \setlength{\tabcolsep}{2pt}
    \renewcommand{\arraystretch}{1.2}
    \footnotesize
     \caption{\textbf{Evaluation of text-to-image generation ability on GenEval~\cite{ghosh2023geneval} benchmark.} $\dagger$ refers to methods using an LLM rewriter.}
    \resizebox{0.8\textwidth}{!}{
    \begin{tabular}{lccccccc}
        \toprule
        \textbf{Model}  & \textbf{Single Obj.} & \textbf{Two Obj.} & \textbf{Counting} & \textbf{Colors} & \textbf{Position} & \textbf{Color Attri.} & \textbf{Overall$\uparrow$} \\
        \midrule
        PixArt-$\alpha$~\cite{chen2024pixart} &  0.98 & 0.50 & 0.44 & 0.80 & 0.08 & 0.07 & 0.48 \\
        Emu$3$-Gen ~\cite{emu3}  & 0.98 & 0.71 & 0.34 & 0.81 & 0.17 & 0.21 & 0.54 \\
        DALL-E $3$~\cite{dalle3} & 0.96 & 0.87 & 0.47 & 0.83 & 0.43 & 0.45 & 0.67 \\
        SD3-Medium~\cite{SD3} & 0.99 & 0.94 & 0.72 & 0.89 & 0.33 & 0.60 & 0.74 \\
        FLUX.1-dev$^{\dagger}$~\cite{flux} & 0.98 & 0.93 & 0.75 & 0.93 & 0.68 & 0.65 & \emph{0.82} \\
        SEED-X~\cite{seed-x}  & 0.97 & 0.58 & 0.26 & 0.80 & 0.19 & 0.14 & 0.49 \\
        Transfusion~\cite{transfusion} & - & - & - & - & - & - & 0.63 \\
        Show-o~\cite{show-o} &  0.98 & 0.80 & 0.66 & 0.84 & 0.31 & 0.50 & 0.68 \\
        Janus-Pro-7B~\cite{januspro2025} &  0.99 & 0.89 & 0.59 & 0.90 & 0.79 & 0.66 & 0.80 \\
        MetaQuery-XL$^{\dagger}$~\cite{pan2025transfer} &  -& - & - & -& -& -& 0.80 \\
        BAGEL$^{\dagger}$~\cite{deng2025emerging} & 0.98 & 0.95  & 0.84 & 0.95 & 0.78 & 0.77 & 0.88 \\
        \ours{}$^{\dagger}$ & 0.99 & 0.97  & 0.78 & 0.83 & 0.74 & 0.65 & 0.83 \\
        \bottomrule
    \end{tabular}
    }
    \label{tab:geneval}
\end{table*}

\subsection{Ablations}
We ablate our design choices in \cref{tab:ablation:clipfid_fid}, ensuring fair and interpretable comparisons and highlighting the effectiveness of our complete \ours{} framework. Comparing the model variant trained solely on internal data with \ours{}, which is trained on the same internal data augmented with our public \dataset{}, we observe consistent performance gains across all generation modes. This demonstrates the effectiveness and value of the newly proposed \dataset{}. To examine the impact of language formulation, we convert the captions into instruction-based prompts. The resulting variant achieves comparable or slightly lower performance than \ours{}, indicating that our framework is robust to different language input formats. We further train three independent models, each dedicated to a single generation mode. Unified \ours{} still achieves superior performance, suggesting that joint training enables beneficial knowledge sharing and synergy across the different generation tasks.

\begin{table*}[t]
    \setlength\tabcolsep{4pt}
    \renewcommand{\arraystretch}{1.1}
    \centering
    \caption{\textbf{Ablation study.} Each cell reports \textbf{CLIP-FID / FID} ($\downarrow$). Training with \dataset{} improves over the internal-only variant, confirming its benefit. The instruction-based setting yields comparable results, showing robustness to language variation. The unified model outperforms single-task models, indicating synergy across generation modes.}
    \scalebox{0.85}{
    \resizebox{\textwidth}{!}{%
    \begin{tabular}{l cc ccc ccc}
        \toprule
        \multirow{2}{*}{Ablation} & \multicolumn{2}{c}{FG\_Gen} & \multicolumn{3}{c}{BG\_Gen} & \multicolumn{3}{c}{Text2All} \\
        \cmidrule(lr){2-3}\cmidrule(lr){4-6}\cmidrule(lr){7-9}
        & Composite & FG & Composite & BG & FG & Composite & BG & FG \\
        \midrule
        Internal data                 & 11.3/79.7 & 28.9/155.9 & 15.9/102.9 & 17.8/134.0 & 22.5/145.2 & 19.6/120.4 & 18.1/136.2 & 28.8/158.0 \\
        Instruction                   & 10.6/77.1 & 27.4/153.1 & 15.2/107.0 & 17.2/134.8 & \textbf{22.6}/140.9 & 18.0/119.5 & \textbf{16.8}/134.3 & \textbf{26.6}/152.5 \\
         Separate task    & 11.0/79.7 & 28.4/156.7 & 15.6/104.1 & 17.8/133.6 & 37.2/142.0 & 19.2/119.6 & 18.4/133.4 & 28.6/153.3 \\
        \ours{}   & \textbf{10.3/75.9} & \textbf{27.3/151.8} & \textbf{14.6/102.9} & \textbf{16.8/132.5} & 22.7/\textbf{139.3} & \textbf{17.8/119.4} & 17.1/\textbf{129.3} & 28.1/\textbf{150.3} \\
        \bottomrule
    \end{tabular}%
    }}
    \label{tab:ablation:clipfid_fid}
\end{table*}

\section{More Visualization Results from \ours{}}
\label{sec:more_model_results}

\subsection{More Qualitative Results from \ours{} under three generation modes}

\begin{figure*}[t]
  \centering
    \includegraphics[width=1.0\linewidth]{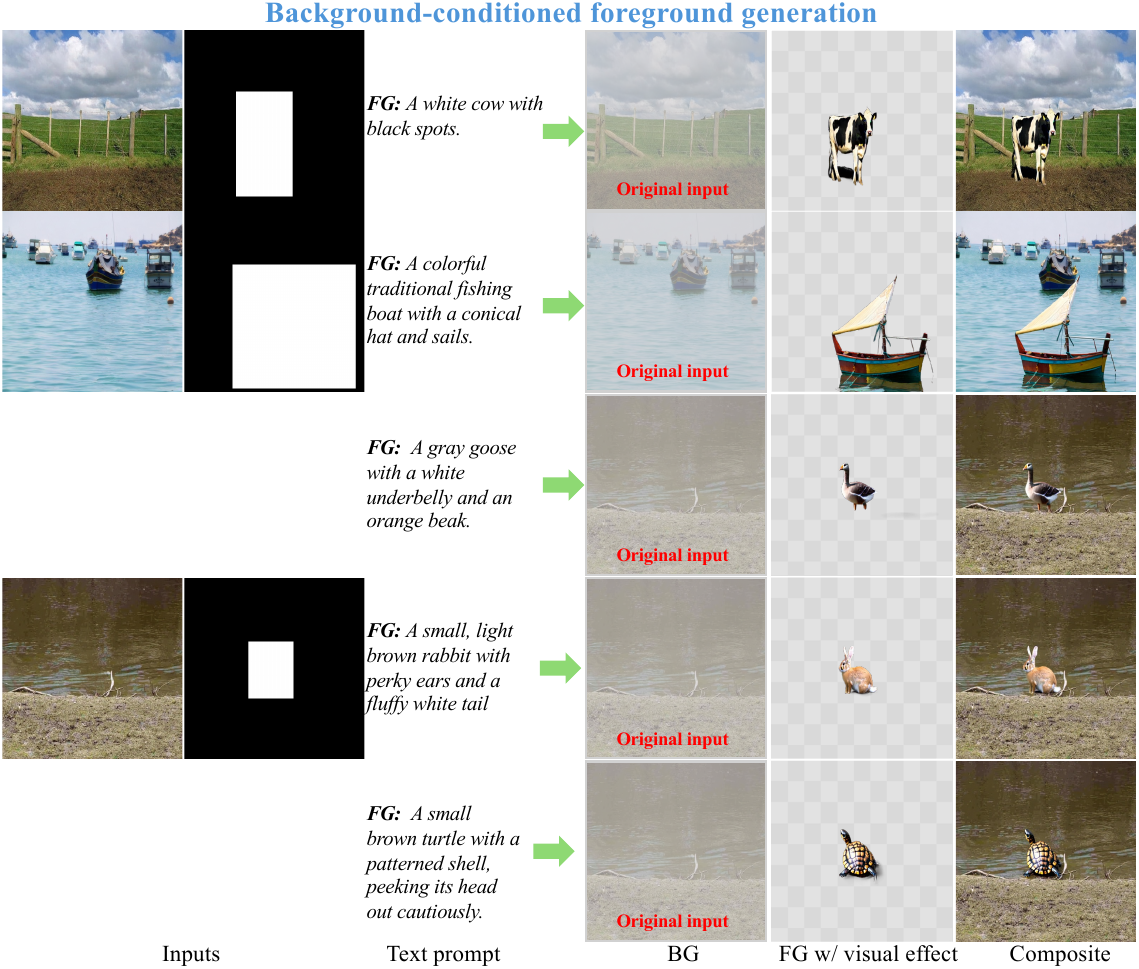}
   \caption{\textbf{More results from \ours{} under the background-conditioned foreground generation (FG\_Gen) mode.}}
   \label{fig:vis_train_1}
\end{figure*}

\begin{figure*}[t]
  \centering
\includegraphics[width=1.0\linewidth]{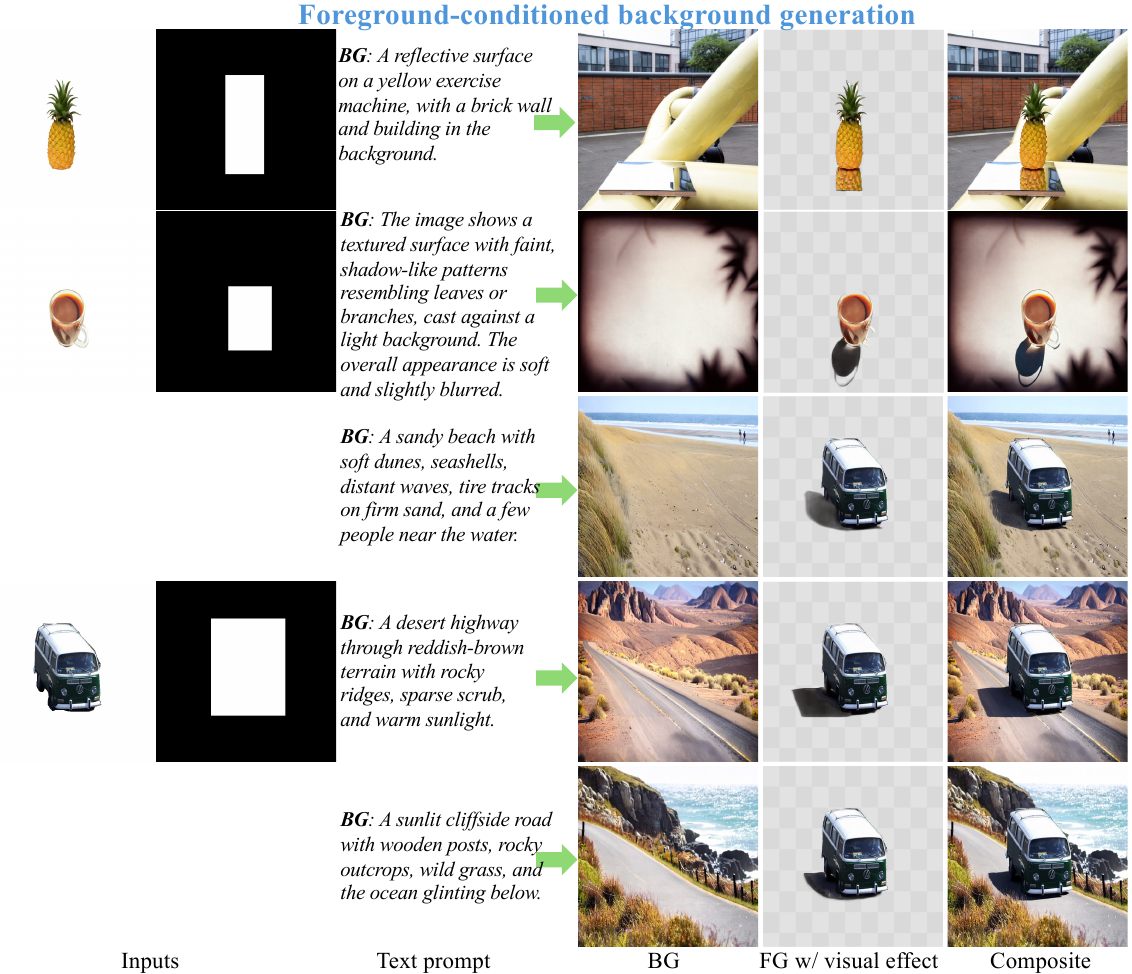}
   \caption{\textbf{More results from \ours{} under the foreground-conditioned background generation (BG\_Gen) mode.}}
   \label{fig:vis_train_2}
\end{figure*}

\begin{figure*}[t]
  \centering
    \includegraphics[width=1.0\linewidth]{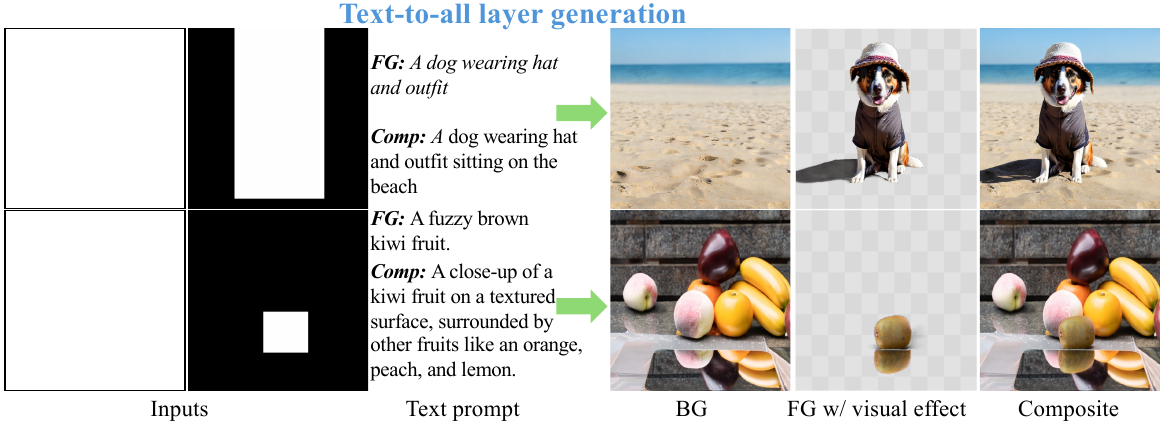}
   \caption{\textbf{More results from \ours{} under the text-to-all layer generation (Text2All) mode.} }
   \label{fig:vis_train_3}
\end{figure*}

As shown in \cref{fig:vis_train_1}, \cref{fig:vis_train_2}, and \cref{fig:vis_train_3}, we provide more results of our model under three different modes (FG\_Gen, BG\_Gen, and Text2All).

Specifically, in \cref{fig:vis_train_1} and \cref{fig:vis_train_2}, for the FG\_Gen and BG\_Gen modes, we further demonstrate more flexible applications. We can fix the background and generate different foregrounds, or fix the foreground and generate different backgrounds.

\subsection{Flexibility of the Foreground-Conditioned Background Generation (BG\_Gen) mode}

\begin{figure*}[t]
  \centering
    \includegraphics[width=1\linewidth]{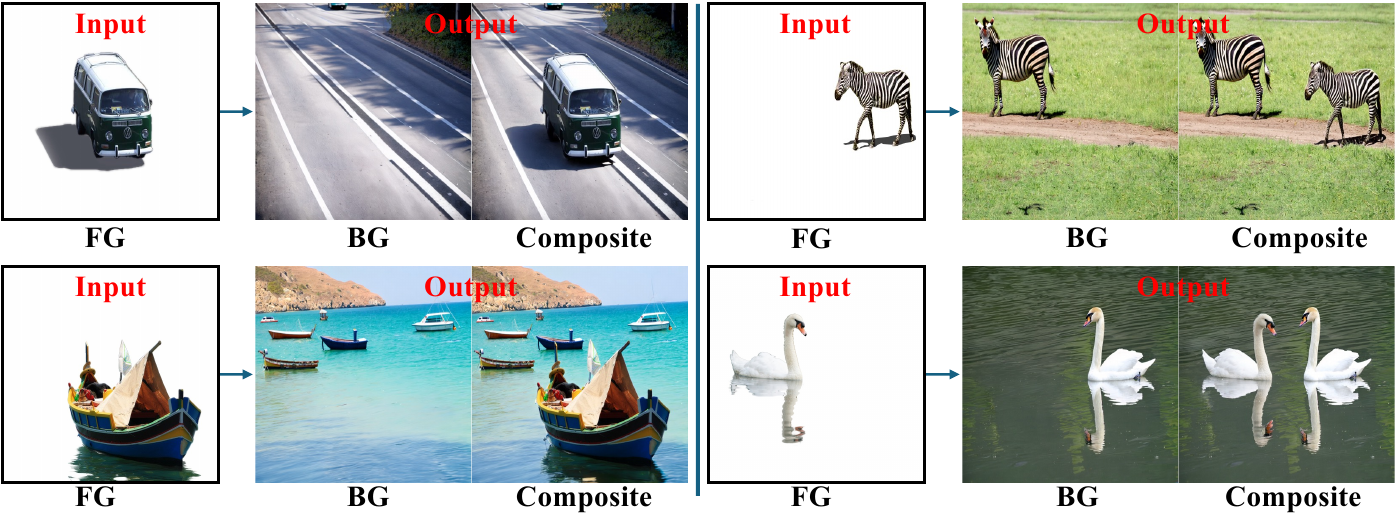}
   \caption{\textbf{Foregrounds with visual effects}. When the input foreground already contains visual effects (e.g., shadows or reflections), our model preserves these effects and generates backgrounds that remain physically and visually consistent with the provided cues.}
   \label{fig:vis_fg_ve}
\end{figure*}

Our primary use case assumes that user-provided foregrounds (FGs) are clean, i.e., without visual effects (VEs). This assumption aligns with most real-world editing workflows, where extracted assets typically lack physically consistent visual effects such as shadows, reflections, or lighting interactions. In this setting, the key advantage of our model is its ability to synthesize the missing effects and generate backgrounds that are physically and visually consistent with the foreground.

Nevertheless, our framework is not limited to this clean-foreground setting. When the input foreground already contains visual effects (e.g., shadows or reflections), our model is able to preserve and respect these existing effects during background generation. As illustrated in Fig.~\ref{fig:vis_fg_ve}, the generated background adapts to the visual cues present in the input foreground, producing lighting and geometric relationships that remain consistent with the provided effects.

This flexibility arises from the fact that our model is trained on real triplets of foreground, visual effects, and background, allowing it to learn the joint distribution between background structures and visual effects. As a result, the model can both synthesize missing effects when they are absent and maintain existing effects when they are present, enabling robust behavior across diverse real-world editing scenarios.

\subsection{More Qualitative Results from Diverse Domains}

\begin{figure*}[t]
  \centering
    \includegraphics[width=1\linewidth]{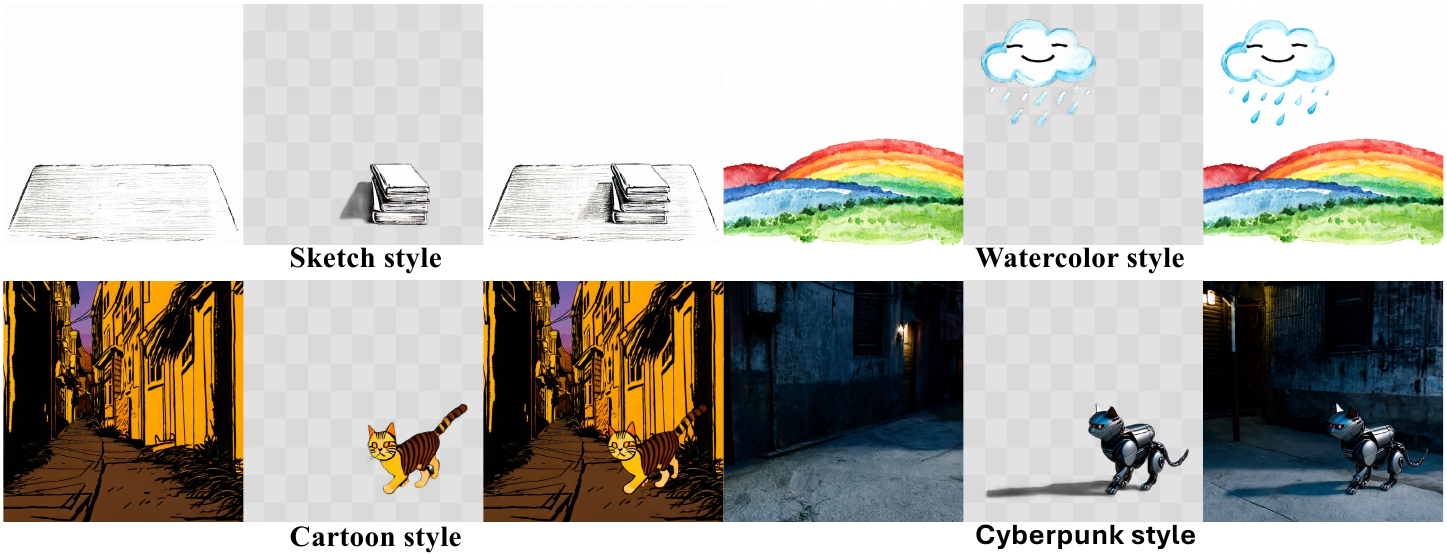}
   \caption{\textbf{More results from \ours{} under the Text-to-All layer generation (Text2All) mode across diverse domains.}}
   \label{fig:vis_domain}
\end{figure*}

As shown in Fig.~\ref{fig:vis_domain}, we further present results on more diverse domains, including \emph{sketch}, \emph{watercolor}, \emph{cartoon}, and \emph{cyberpunk} styles. Although our \dataset{} dataset is composed of real-world imagery, our method performs well even in these stylistically distinct domains. We attribute this generalization ability to the physical knowledge embedded in our real-world training data, which enables the model to effectively transfer the knowledge learned by the pretrained model to other domains.

\section{More Samples from \dataset{} and \benchmark{}}
\label{sec:more_samples_dataset}

\benchmark{} comes from 6 different data sources, with the specific quantity distribution shown in Table~\ref{tab:benchmark_distri}.

\begin{table}[t]
    \centering
    \small
    \caption{\textbf{Benchmark statistics.} \benchmark{} is built from 6 distinct sources---four public datasets~\cite{tudosiu2024mulan,lin2014microsoft,wang2022instance,unsplash} and two in-house data---to ensure diversity and representativeness.}
    \resizebox{0.75\columnwidth}{!}{%
    \begin{tabular}{l cccccc}
        \toprule
        \textbf{Data Source} & \textbf{MULAN} & \textbf{COCO 2017} & \textbf{SOBA} & \textbf{Unsplash} & \textbf{Camera-Indoor} & \textbf{Camera-Outdoor} \\
        \midrule
        \textbf{Num. Images} & 45 & 40 & 50 & 27 & 40 & 40 \\
        \bottomrule
    \end{tabular}%
    }
\label{tab:benchmark_distri}
\end{table}

\begin{figure*}[t]
  \centering
    \includegraphics[width=1\linewidth]{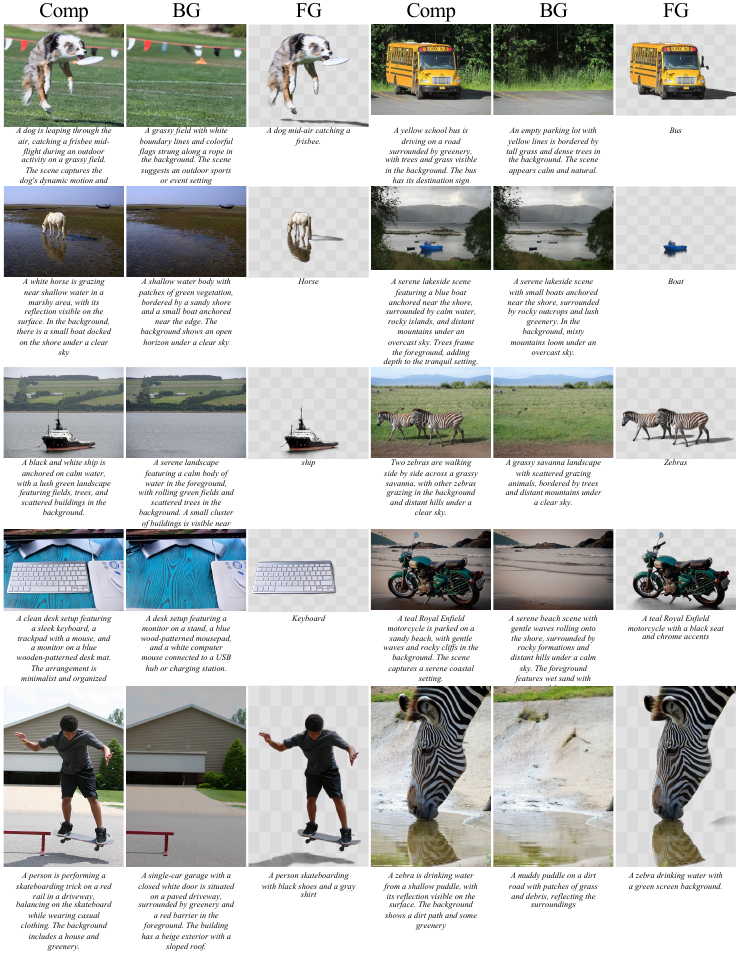}
   \caption{\textbf{More samples of \dataset{}.} Each  sample consists of a composite image, a clean background, and a foreground layer with visual effects, along with corresponding captions for all components. }
   \label{fig:vis_train_data}
\end{figure*}

As shown in \cref{fig:vis_train_data}, we provide more samples from \dataset{}.

\begin{figure*}[t]
  \centering
    \includegraphics[width=1\linewidth]{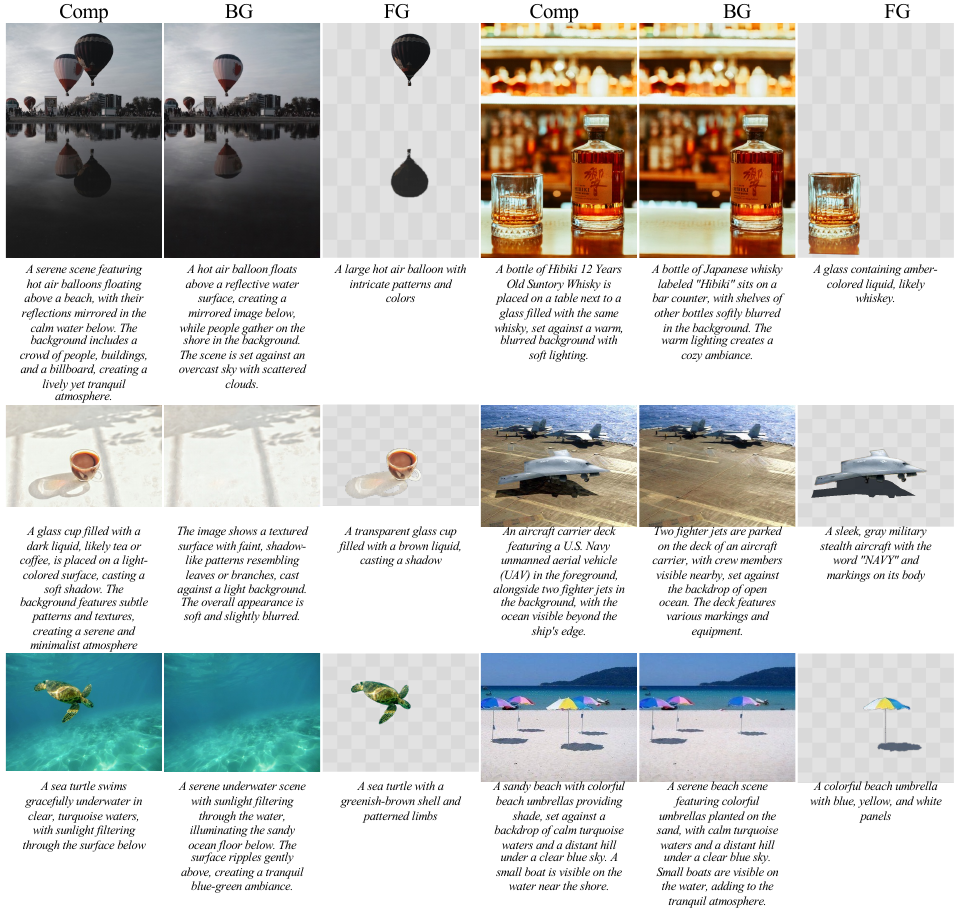}
   \caption{\textbf{More samples of \benchmark{}.} Each  sample consists of a composite image, a foreground layer with visual effects, and a clean background along with corresponding captions for all components. \textbf{Foreground with visual effects annotated by professional annotators. Background are decomposed by expert models or captured by camera.} }
   \label{fig:vis_benchmark_data}
\end{figure*}

As shown in \cref{fig:vis_benchmark_data}, we provide more samples from \benchmark{}.

\section{Detailed Scores of the GPT Score}

As shown in \cref{tab:quant:clipfid_fid} in the main manuscript, we provide the ``GPT Score'', which is the average score of the instruction-following and identity-preserving metrics proposed by Complex-Edit~\cite{yang2025complexedit}. Here, we additionally provide the original score in \cref{tab:gpt_score}. We run the same prompt three times and take the average as the original score for instruction-following and identity-preserving.

\begin{table*}[t]
    \centering
    \setlength{\tabcolsep}{4pt}
    \renewcommand{\arraystretch}{1.2}
    \small
    \caption{\textbf{GPT scores.} ``FG\_Gen'' denotes background-conditioned foreground layer generation, ``BG\_Gen'' denotes foreground-conditioned background generation, and ``Text2All'' denotes text-to-all layer generation. ``IF'' stands for Instruction Following and ``IP'' for Identity Preservation. Results for models marked with $\ast$ are obtained using their respective expert models rather than a single unified model. Specifically, for the FG\_Gen and BG\_Gen tasks, we use the FLUX.1-Fill-dev, Qwen-Image-Edit-2509, and gpt-image-1[high] editing models, respectively. For the All\_Gen task, we use the FLUX.1-schnell, Qwen-Image, and gpt-image-1[high] models as text-to-image models, respectively.}
    \begin{tabular}{lcccccccccccc}
        \toprule
        \multirow{2}{*}{Model}
        & \multicolumn{3}{c}{FG\_Gen} & & \multicolumn{3}{c}{BG\_Gen} & & \multicolumn{3}{c}{Text2All} \\
        \cmidrule(lr){2-4} \cmidrule(lr){6-8} \cmidrule(lr){10-12}
        & IF $\uparrow$ & IP $\uparrow$ & Avg
        &
        & IF $\uparrow$ & IP $\uparrow$ & Avg
        &
        & IF $\uparrow$ & IP $\uparrow$ & Avg \\
        \midrule
        gpt-image-1[high]\textsuperscript{$\ast$}  & 9.77 & 7.88 & 8.8  && 9.84 & 7.95 & 8.9  && 6.95 & 7.26 & 7.1  \\
        FLUX.1\textsuperscript{$\ast$}             & 8.32 & 9.46 & 8.9  && 8.43 & 8.48 & 8.5  && 6.46 & 5.61 & 6.0  \\
        Qwen-Image-Edit\textsuperscript{$\ast$}    & 8.66 & 9.34 & 9.0  && 7.25 & 8.46 & 7.9  && 6.41 & 4.98 & 5.7  \\
        \textbf{Ours}                              & 9.27 & 9.29 & 9.3  && 9.34 & 8.67 & 9.0  && 8.31 & 6.91 & 7.6  \\
        \bottomrule
    \end{tabular}
    \label{tab:gpt_score}
\end{table*}

\section{Details of the Metric in  \cref{sec:ve_necessity}  (The Necessity of Layered Generation with Visual Effects)}
\label{sec:ve_necessity_supp}
  
To illustrate the necessity of our proposed generation paradigm (Layer Editing with Visual Effects), we conduct quantitative experiments in \cref{sec:ve_necessity} of the main manuscript. The results show the superiority of our generation paradigm.

We compare three editing modes: \emph{Instruct Editing}, \emph{Cascaded Layer Editing}, and \emph{Layer Editing with Visual Effects}.
For both \emph{Cascaded Layer Editing} and \emph{Layer Editing with Visual Effects}, the foreground is represented in RGBA format, which allows us to perform programmatic, pixel-accurate modifications. This also ensures that the editing parameters remain fully consistent across the two modes.

We benchmark these three approaches on recoloring, spatial editing, and complex compositional editing tasks. For recoloring, we define seven random color transformation operations. For spatial editing, we randomly select a movement direction (up, down, left, or right) and apply one of three displacement magnitudes ($20\%$, $30\%$, or $50\%$). For compositional editing, we randomly combine recoloring and spatial editing to test multi-factor control. All evaluations are conducted automatically to ensure objective and reproducible comparison across methods.

\begin{figure*}[t]
  \centering
    \includegraphics[width=1\linewidth]{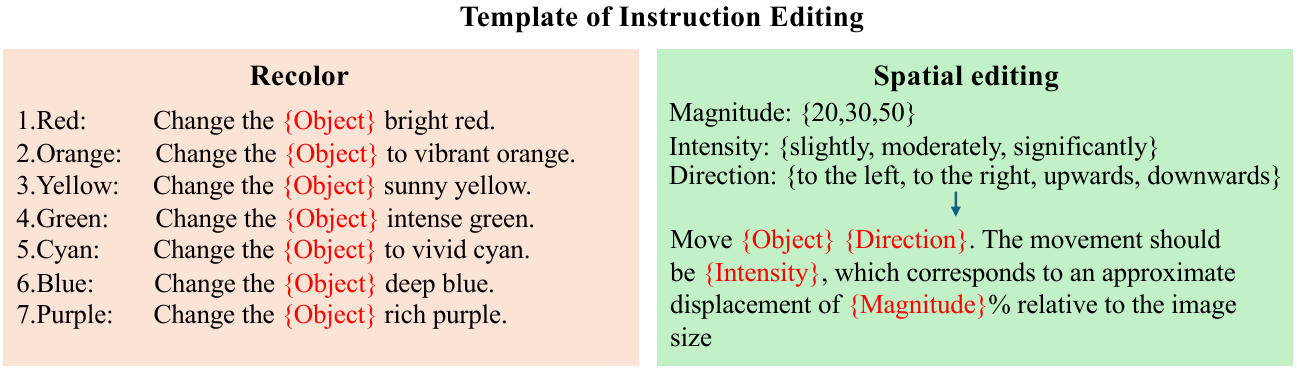}
   \caption{When performing instruct editing, the content within \textbf{\textcolor{red}{\{\}}}  will be replaced by specific parameters. In spatial editing, the values of the Magnitude and Intensity parameters correspond one-to-one from left to right.}
   \label{fig:vis_instruct}
\end{figure*}

For \emph{Instruct Editing} mode, when performing recolor and spatial editing, we input the editing parameters from the previous Layer Editing into GPT-5. Based on the context of our question and the value of the parameter, GPT-5 generates an appropriate natural language description, as shown in the template in \cref{fig:vis_instruct}. In the subsequent Complex Editing task, we combine the two types of instructions accordingly. This helps ensure that all three types of editing perform the same actions as much as possible, thereby ensuring comparability.

\section{Training Samples of Data Curator}

\begin{figure*}[t]
  \centering
    \includegraphics[width=1\linewidth]{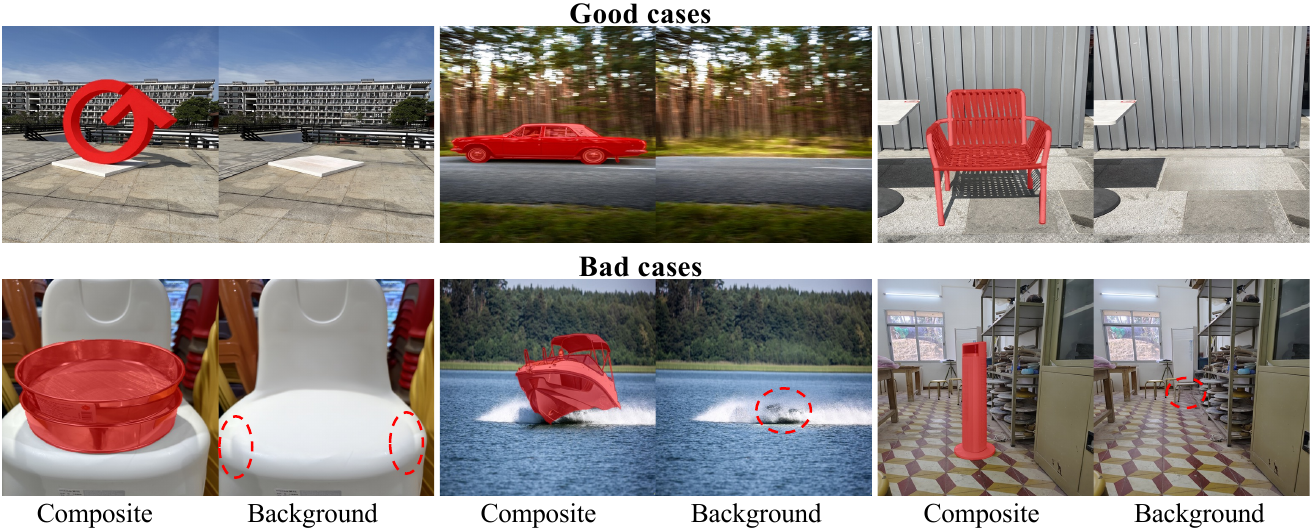}
   \caption{\textbf{Training Samples of Data Curator.} The red dashed box indicates the main problematic area in the bad cases.}
   \label{fig:data_curator_train_data}
\end{figure*}

In \cref{sec:LASAGNA-48K}, \dataset{} Dataset in the main manuscript, we demonstrate the complete data construction pipeline. The data curator is a key component in the data construction pipeline. To obtain high-quality data filtering results, we carefully annotated about 30K samples by humans, as shown in \cref{fig:data_curator_train_data}, to train the data curator. Each sample is a triplet data: a composite image, mask and a background without the foreground, which are input into the data curator. The curator outputs a confidence score (between 0 and 1) indicating the probability that the background is good.

In manual annotation, we provide a binary label for each data sample. Specifically, we determine the final binary label based on the following three aspects:

\begin{itemize}
    \item \textbf{Hole filling}: Whether the object has been successfully removed.
    \item \textbf{Background consistency}: Whether the removed area is consistent with the surrounding environment.
    \item \textbf{Visual effect}: Whether the visual effects caused by the object (\eg, shadow and reflection) have been removed.
\end{itemize}

If \textbf{any one} of the above criteria is not met, the sample is considered a bad case. Only when all conditions are satisfied is it labeled as a good case. We highlight these unnatural regions with dashed boxes in \cref{fig:data_curator_train_data}. These hard negative samples ensure that, after training the data curator, the filtered data it produces consists exclusively of high-quality background images.

\end{document}